%% file: main.tex
\documentclass[acmsmall]{acmart}
\AtBeginDocument{%
  \providecommand\BibTeX{{%
    \normalfont B\kern-0.5em{\scshape i\kern-0.25em b}\kern-0.8em\TeX}}}

\setcopyright{acmcopyright}
\copyrightyear{2023}
\acmYear{2023}
\acmDOI{10.1145/3626312}

\acmJournal{CSUR}
\acmArticle{1}
\acmMonth{10}

\usepackage{makecell}
\usepackage{subfigure}
\usepackage{bm}
\usepackage{multirow}
\usepackage{multicol}
\usepackage{tabularx}
\usepackage{booktabs}
\usepackage{bbding}
\usepackage{dsfont}
\usepackage{mathrsfs,amsmath}
\usepackage{mathrsfs}

\begin{document}

\title{A Survey of Deep Learning for Low-Shot Object Detection}


\author{Qihan Huang}
\affiliation{%
  \institution{Zhejiang University}
  \country{China}
  }
  
\author{Haofei Zhang}
\affiliation{%
  \institution{Zhejiang University}
  \country{China}}
  
\author{Mengqi Xue}
\affiliation{%
  \institution{Zhejiang University}
  \country{China}
  }
  
\author{Jie Song}
\affiliation{%
  \institution{Zhejiang University}
  \country{China}
  }
  
\author{Mingli Song${}^{\dagger}$}
\affiliation{%
  \institution{Zhejiang University}
  \country{China}
  }

\renewcommand{\shortauthors}{Huang, et al.}

\footnotetext[1]{$\dagger$ Corresponding author.}%
\footnotetext[2]{This work is funded by National Natural Science Foundation of
   China (U20B2066, 61976186, 62106220), Ningbo Natural Science Foundation
   (2021J189), and the Fundamental Research Funds for the Central
   Universities (2021FZZX001-23, 226-2023-00048).}

\begin{abstract}
Object detection has achieved a huge breakthrough with deep neural networks and massive annotated data. However, current detection methods cannot be directly transferred to the scenario where the annotated data is scarce due to the severe overfitting problem. Although few-shot learning and zero-shot learning have been extensively explored in the field of image classification, it is indispensable to design new methods for object detection in the data-scarce scenario since object detection has an additional challenging localization task.
Low-Shot Object Detection~(LSOD) is an emerging research topic of detecting objects from a few or even no annotated samples, consisting of One-Shot Object Localization~(OSOL), Few-Shot Object Detection~(FSOD), and Zero-Shot Object Detection~(ZSOD).
This survey provides a comprehensive review of LSOD methods. First, we propose a thorough taxonomy of LSOD methods and analyze them systematically, comprising some extensional topics of LSOD~(semi-supervised LSOD, weakly-supervised LSOD, and incremental LSOD).
Then, we indicate the pros and cons of current LSOD methods with a comparison of their performance.
Finally, we discuss the challenges and promising directions of LSOD to provide guidance for future works.
\end{abstract}

\ccsdesc[500]{General and reference~Surveys and overviews}
\ccsdesc[500]{Computing methodologies~Object detection}

\keywords{Few-Shot Object Detection, One-Shot Object Detection, Zero-Shot Object detection, Transfer-Learning, Meta-Learning}

\maketitle

\input{n00_Introduction}
\input{n01_Preliminaries}
\input{n02_OSOL}
\input{n03_FSOD}
\input{n05_ZSOD}
\input{n06_benchmarks}
\input{n07_directions}
\input{n08_conclusions}

\bibliographystyle{ACM-Reference-Format}
\bibliography{references}


\end{document}

%% file: n00_Introduction.tex
\section{Introduction}
Object detection is a fundamental yet challenging task in computer vision, aiming to locate objects of certain classes in images. It has been widely applied to many computer vision tasks like object tracking \cite{alper2006tracking, wang2021FOunify, Paul2019MOTS}, autonomous driving \cite{Sorin2020auto, Ekim2020auto}, scene graph generation \cite{yao2021graph, Yang2018graphrcnn, Tang2020sgg}.

The general process of object detection is to predict classes for a set of bounding boxes (imaginary rectangles for reference in the image). Most traditional methods are slow since they generate the bounding boxes using brute force by sliding a window through the whole image. Viola-Jones (VJ) detector \cite{paul2001VJ} first achieves real-time detection of human faces with three speed-up techniques: integral image, feature selection, and detection cascades. Later, histogram of oriented gradients (HOG)~\cite{2005DalalHOG} is proposed, and many traditional object detectors adopt it for feature description. Deformable part-based model (DPM) \cite{Pedro2008DPM} is a representative traditional method. DPM divides an object detection task into several fine-grained detection tasks, then uses some part-filters to detect parts of an object and aggregates them for final prediction. Although people have made many improvements, traditional methods are restricted by their slow speed and low accuracy.

\input{figures/LSOD_category}

Compared with these traditional methods, deep-learning-based methods have significantly improved performance. Current deep detectors roughly consist of two-stage detectors and single-stage detectors. Two-stage detectors first generate region proposals (i.e., image regions which are more likely to contain objects) and next make predictions on them, following a similar framework to traditional methods.
R-CNN \cite{Ross2014RCNN} is one of the earliest works of two-stage detectors. It uses selective search to obtain region proposals then extracts their features with a pre-trained CNN model for further classification and regression. Fast R-CNN \cite{Girshick_2015_ICCV} improves R-CNN by using a region of interest (RoI) pooling layer to generate feature maps for region proposals from the integral feature map. Faster R-CNN \cite{ren2015faster_rcnn} further proposes a region proposal network (RPN) to generate region proposals from the whole image feature map using anchors (i.e., pre-defined bounding boxes with specific height and width).
However, the generation of region proposals requires high computation cost and storage costs. To mitigate this problem, single-stage detectors are proposed to combine these two stages.
YOLO-style object detectors \cite{redmon2018yolov3, alexey2020yolov4, zheng2021yolox} are the representative works of single-stage detectors. Given the feature map extracted from the original image, YOLO-style detectors directly pre-define anchors with multiple scales over all locations of the image and predict the class probabilities, regression offsets and object confidence scores of each anchor. Single-stage detectors achieve higher speed, but they generally underperform two-stage detectors. Moreover, some methods like focal loss \cite{lin2017focal} have been proposed to decrease the performance gap between single-stage and two-stage detectors.
Recently, a transformer-based detector named DETR \cite{NicolasDETR2020} has been proposed. DETR achieves end-to-end detection and has comparable performance to many classic detectors. Some extended methods \cite{zhu2021deformable, dai2020updetr} are proposed to mitigate the slow convergence problem of DETR.

However, these deep detectors tend to overfit when the training data is scarce and thus require abundant annotated data.
In real life, it is hard to collect sufficient annotated data for some object classes due to the scarcity of these classes or special labeling costs, and current deep detectors are not competent in this situation.
Therefore, the ability to detect objects from a few or even zero annotated samples is desired for modern detectors. To achieve this goal, Low-Shot Object Detection (LSOD) is introduced into object detection, including One-Shot Object Localization (OSOL), Few-Shot Object Detection (FSOD), Zero-Shot Object Detection (ZSOD). These three settings of LSOD mainly differ in the number of annotated samples for each category. Concretely, OSOL and FSOD tackle the situation that each object category has one or more annotated image samples, while ZSOD differentiates different classes according to the semantic information of each category instead of image samples.

OSOL and FSOD are developed following the mainstream scheme of few-shot learning~(FSL). Few-shot learning divides the object classes into base classes with many annotated samples (denoted as base dataset) and novel classes with a few annotated samples (denoted as novel dataset).
Note that the annotated samples and the test samples in novel classes are named as support samples and query samples, respectively.
Few-shot learning requires to pre-train the model on the base dataset then uses the model to predict novel classes on the novel dataset for evaluation.
Current few-shot learning methods are roughly categorized into meta-learning methods and transfer-learning methods.
Meta-learning methods adopt a ``learning-to-learn'' mechanism, which defines multiple few-shot tasks on the base dataset to train the model, and enables the model to adapt to the real few-shot tasks quickly. Moreover, transfer-learning methods learn a good image representation by directly training the model on the base dataset, which is used for the novel dataset. Although meta-learning is a more natural approach to tackle the few-shot problem, Tian et al. \cite{tian2020rethinking} find that the baseline transfer-learning methods surpass some classic meta-learning methods, especially in the cross-domain few-shot learning.
Current few-shot learning methods are mainly explored on the task of image classification. OSOL and FSOD are more challenging than few-shot image classification because object detection requires an extra task to locate the objects. As the branches of few-shot learning, OSOL and FSOD also inherit the core methods (meta-learning \& transfer-learning) of it.

OSOL is a few-shot learning setting on object detection which locates objects using only one labeled image of each category in the image. Current OSOL methods all adopt the scheme of meta-learning following few-shot learning, where a large number of one-shot tasks are defined on the base dataset to train the model. OSOL has a strong guarantee that the model precisely knows the object classes contained in each test image. With this strong guarantee, the latest OSOL methods have achieved relatively high performance.

\input{tables/notations}

However, OSOL setting is not realistic enough since the object classes in the test images are not pre-known in real life. Therefore, another few-shot setting on object detection is adopted by more papers, which is named Few-Shot Object Detection~(FSOD). The major differences between FSOD and OSOL are as follows: (1) FSOD needs to predict the correct category of potential objects in the test image. (2) OSOL samples support images independently for each test image, FSOD samples the support images only once for all test images. (3) In FSOD, the number of labeled samples per category can be larger than one. Similar to methods on few-shot image classification, FSOD methods are categorized into two mainstream methods: meta-learning methods and transfer-learning methods. Early FSOD methods mainly adopt the meta-learning scheme.
The core operation of meta-learning FSOD methods is to extract the features of a few annotated samples (support features) and aggregate them into the features of query images~(query features) for guidance on the prediction of query images.
This aggregation operation promotes the model to learn adequate information from a few annotated samples. Early meta-learning FSOD methods simply aggregate the support features with the features of RoIs~(RoI features) in the query images. Afterwards, researchers find that the aggregation of integral features is essential for performance improvement since the shallow components in the model also require the information of annotated samples~(e.g., the RPN component in Faster R-CNN needs the support features to filter out unmatched region proposals).
Therefore, this survey categorizes meta-learning FSOD methods into RoI feature aggregation and mixed feature aggregation methods~(``mixed'' means using RoI feature aggregation and integral feature aggregation together).
Unlike meta-learning methods, transfer-learning FSOD methods directly pre-train the detector on the base dataset and fine-tune it on the novel dataset.
Early FSOD methods rarely adopt transfer-learning due to its poor performance. TFA~\cite{wang2020few} subverts this cognition, which proposes a two-stage fine-tuning strategy to fine-tune the model and achieves better performance than contemporary meta-learning methods.
In addition to the standard FSOD discussed above, other extensional settings like semi-supervised FSOD~\cite{MisraSHWatchAndLearn15, dong2018few}, weakly-supervised FSOD~\cite{jiyanggaoNOTERCNN2019, LeonidStarNet2021} and incremental FSOD~\cite{PerezRuaZHXIncremental2020, pengyangliClassIncremental2021} are also explored by researchers and investigated in this survey.

ZSOD assigns abundant labeled samples to base classes, but it assigns no annotated image samples to novel classes. Instead, mainstream ZSOD allocates semantic attributes to each class (including base and novel classes), and it classifies object proposals according to their semantic similarities with different classes.
Mainstream ZSOD methods include visual-semantic mapping methods, semantic relation methods, and data augmentation methods.
Most early ZSOD methods belong to visual-semantic mapping methods. These methods aim to learn a visual-semantic function using the annotated samples of the base dataset, which projects visual features into semantic embeddings for comparison with class semantic attributes. Next, semantic relation methods utilize the semantic relation between different classes to make predictions. Moreover, data augmentation methods attempt to generate visual samples for novel classes and re-train the model.
Besides the mainstream ZSOD setting described above, this survey discusses some rarely explored settings like transductive ZSOD and textual-description-based inductive ZSOD.
Recently, with the emergence of large-scale cross-modal models~(e.g., CLIP~\cite{Radford2021ICML}), Open-Vocabulary Object Detection~(OVD) attracts more and more research interest, which first trains a stronger visual-semantic mapping function for multiple classes and significantly improves the performance of the further ZSOD task.

The overview of this survey is illustrated in \autoref{fig:taxonomy}. The preliminaries for \textbf{meta-learning} and \textbf{transfer-learning} are given in \autoref{sec:prelimilaries}. The more fine-grained categorization and analysis of methods for LSOD are described in \autoref{sec:osod_methods}, \autoref{sec:standard_fsod}, \autoref{sec:standard_zsd_methods}, \autoref{sec:non_standard_zsd_methods}. The two popular datasets (MS COCO dataset \cite{Lin2014MsCOCO} and PASCAL VOC dataset \cite{Mark2010Pascal}) and evaluation criteria of LSOD are described in \autoref{sec:dataset_eval}. The performance of current LSOD methods is summarized in \autoref{sec:performance}. The promising directions LSOD are discussed in \autoref{sec:future_dirs}. Finally, \autoref{sec:conclusion} concludes the contents of this survey. The key notations used in this survey are summarized in \autoref{tab:key_notations}.

%% file: figures/LSOD_category.tex
\begin{figure}
\centering
\includegraphics[width=\textwidth]{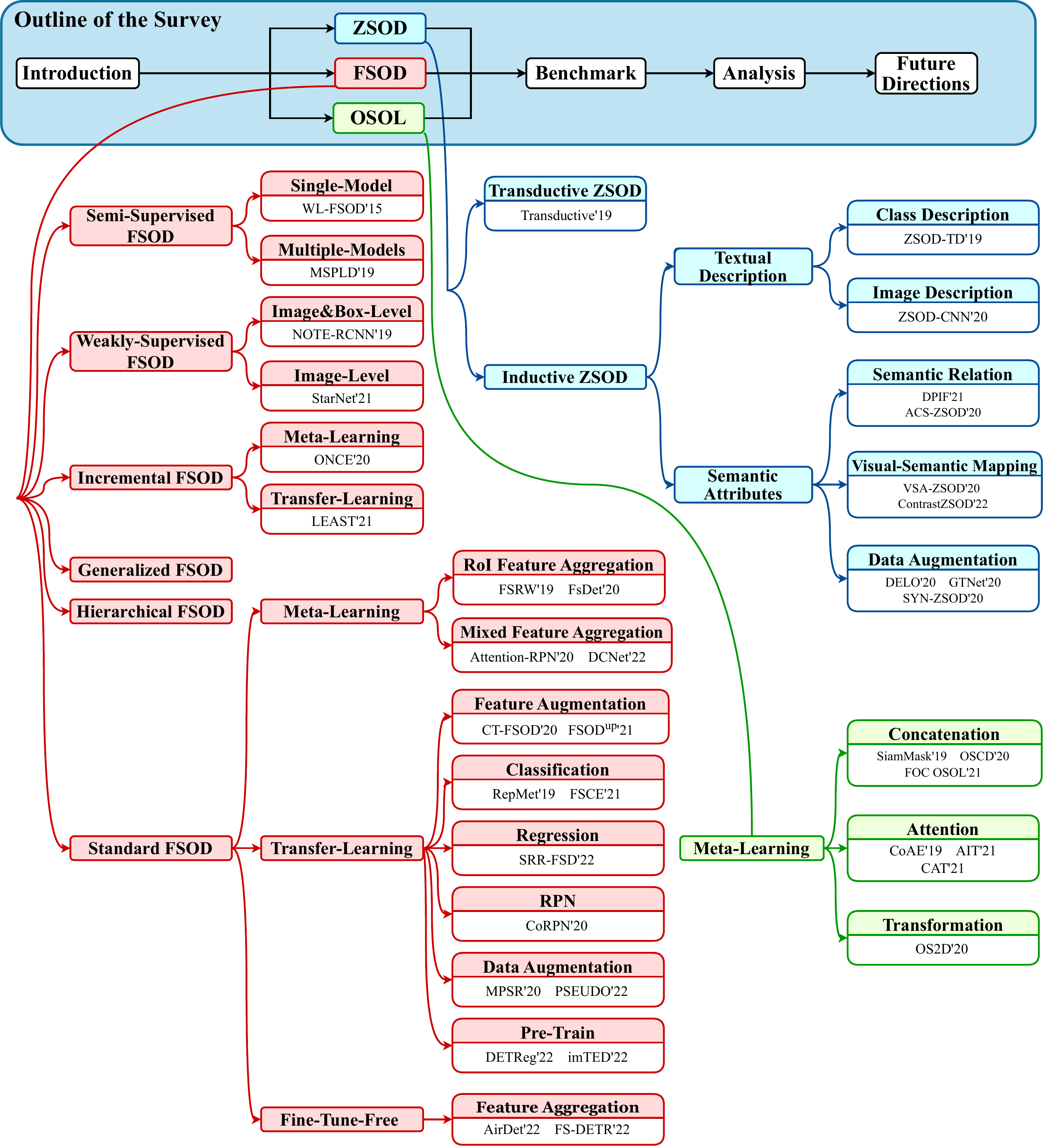}\\
\caption{Overview of this survey. This survey gives a general introduction to Low-Shot Object Detection (LSOD), then categorizes LSOD into three domains: One-Shot Object Localization (OSOL), Few-Shot Object Detection (FSOD) and Zero-Shot Object Detection (ZSOD). The more fine-grained categorization of these three domains is also demonstrated in the figure with three colors, which will be discussed detailedly in later sections. Each category is demonstrated with \textbf{a part of representative works} in the figure. Then the benchmarks for OSOL, FSOD and ZSOD are summarized, and the performance of different LSOD methods on these benchmarks is compared and analyzed. Finally, the future directions of LSOD are discussed.}
\label{fig:taxonomy}
\end{figure}

%% file: tables/notations.tex
\begin{table*}
\footnotesize
\renewcommand\arraystretch{1.3}
\centering
\caption{Key Notations Used in This Article}
\label{tab:key_notations}
\setlength{\tabcolsep}{1.4mm}{
\begin{tabular}{c|l|c|l} 
    \toprule
  \textbf{Notation} & \makecell[c]{\textbf{Description}} & \textbf{Notation} & \makecell[c]{\textbf{Description}}
  \\
  \midrule
  $\phi_q$ & Feature map of integral query image $q$ & ${\rm Pool}(\cdot)$ & Pool operation\\
  $\phi_s$ & Feature map of integral support image & $\oplus$ & Element-wise sum\\
  $\phi_{\rm fused}$ & The aggregated feature map of $\phi_q$ and $\phi_c$ & $\otimes$ & Channel-wise multiplication \\
  $\phi_{r}$ & The RoI feature map in the query image & ${\rm Conv}(\cdot)$ & Convolutional operation \\
  $v_{r}$ & The RoI feature vector in the query image & ${\rm FC}(\cdot)$ & FC layer \\
  $s_{r}$ & The RoI semantic embedding in the query image & ${\rm Softmax}(\cdot)$ & Softmax operation \\ 
  $v_s$ & The pooled feature vector & $\sigma(\cdot)$ & Sigmoid function \\
  $v_{\rm fused}$ & The aggregated feature vector of $v_{q}^{i}$ and $v_c$ & ${\rm RELU}(\cdot)$ & RELU function \\
  $s_c$ & The semantic embedding of class $c$ & $||\cdot||$ & The norm of a vector \\
  $p_c$ & The prediction score for class $c$ of a RoI & $[ \cdot ]$ & Concatenation operation \\
  $|\cdot|$ & The absolute value of a vector & & \\
  \bottomrule
  \end{tabular}}
\end{table*}

%% file: n01_Preliminaries.tex
\section{Preliminaries\label{sec:prelimilaries} \label{sec:preliminaries}}

\input{figures/meta_transfer_illustration}

\subsection{Meta-Learning}

Meta-learning is a ``learning-to-learn''~\cite{thrun1998learning, timothy2022TPAMI} paradigm extended from the conventional ``learning'' paradigm. Conventional learning paradigm directly trains the model from scratch on the whole dataset as a single task. Differently, meta-learning learns the training pattern~(e.g., parameter initialization) from multiple tasks, which is capable of generalizing across different tasks and facilitating the learning of new tasks. Therefore, meta-learning is suitable for quick adaptation of the model to the new tasks in few-shot learning. The framework of meta-learning is shown in \autoref{fig:meta_transfer_illustration}~(a), and a more detailed illustration is in section {\bf S1} of the supplementary online-only material.

\subsection{Transfer-Learning}

Transfer-learning methods aim to transfer the knowledge~(good feature representation) from a related domain~(named source domain) to the current domain~(named target domain), in order to improve the performance of model on the target domain, as shown in \autoref{fig:meta_transfer_illustration}~(b). Traditional transfer-learning approaches include instance-based methods, feature-based methods, parameter-based methods, and relational-based methods~\cite{zhuang2021IEEE}.

For the transfer-learning methods in few-shot learning, the base dataset is viewed as the source domain, and the novel dataset is viewed as the target dataset.
Tian et al.~\cite{tian2020rethinking} find that simply transferring a strong feature extractor from the base dataset to the novel dataset outperforms many meta-learning methods on few-shot image classification, and many FSL methods follow this paradigm.
Transfer-learning is not suitable for OSOL since the target domain consists of only one image for each task, yet it is widely adopted in FSOD after the emergence of TFA~\cite{wang2020few}.

%% file: figures/meta_transfer_illustration.tex
\begin{figure}
\centering
\includegraphics[width=0.85\textwidth]{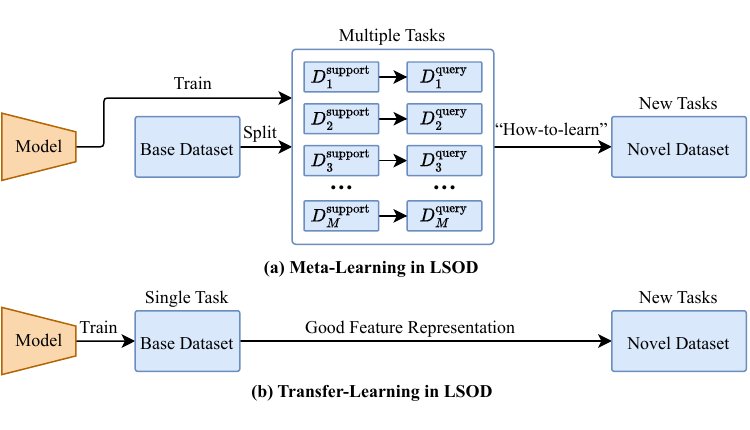}\\
\caption{Illustration of \textbf{meta-learning} and \textbf{transfer-learning} in LSOD. LSOD divides the object classes into base classes with many annotated samples~(denoted as base dataset) and novel classes with a few annotated samples~(denoted as novel dataset). Meta-learning samples multiple tasks from the base dataset and trains the model on these tasks~(each task requires to make predictions on $D_i^{\mathrm{query}}$ according to the annotated $D_i^{\mathrm{support}}$), aiming to acquire the knowledge about ``how to learn'' and generalize it to the novel dataset. On the other hand, transfer-learning directly trains the model on the base dataset and transfers a good feature representation to the novel dataset, enabling the representation of objects from novel classes.}
\label{fig:meta_transfer_illustration}
\end{figure}

%% file: n02_OSOL.tex
\section{One-Shot Object Localization\label{sec:osod_methods}}

\textbf{Task Settings.} One-Shot Object Localization~(OSOL) needs to locate objects in a query image according to only one support image for each novel class existing in this query image. The training dataset~(base dataset $D_B$) of OSOL comprises abundant annotated instances of base classes $C_B$, and the test dataset~(novel dataset $D_N$) comprises instances of novel classes $C_N$~($C_B$ and $C_N$ are not intersected).
Specifically, for each query image in $D_N$, OSOL randomly samples a support image for each novel class existing in the image. Next, OSOL locates the novel objects in the query image according to the corresponding support image.
The main difference from FSOD is that OSOL only requires a binary classification task to discriminate whether the potential object is foreground or background according to the given support image, while FSOD requires a multi-class classification task because FSOD doesn't pre-know the classes of the existing objects in the query images.

\noindent \textbf{Framework of Current OSOL Methods.} Some previous object tracking methods like SiamFC~\cite{Cen2018SiamFC}, SiamRPN~\cite{Li2018SiamRPN} are forerunners of OSOL, which are used for comparison with early OSOL methods.
Current OSOL methods adopt the meta-learning scheme, and their framework is based on Faster R-CNN, as shown in \autoref{fig:osol_framework}.
First, they extract the integral features of the query image and the support image using the same convolutional backbone~(named query features and support features, respectively), then conduct ``integral feature aggregation'' to generate a fused feature map by aggregating the query features with the support features. This fused feature map is fed into RPN and RoI layer to generate category-specific region proposals and the corresponding RoI features, respectively. Finally, these RoI features are used for the final classification and localization tasks. Furthermore, some methods additionally conduct ``RoI feature aggregation'' to further aggregate the RoI features with the support features.

Current OSOL methods mainly differ in the feature aggregation method, and this survey accordingly categorizes OSOL methods into concatenation-based methods, attention-based methods, and transformation-based methods. In the following sections, $\phi_q \in \mathbb{R}^{C \times H_q \times H_q}$, $\phi_r \in \mathbb{R}^{C \times H_r \times H_r}$ and $\phi_s \in \mathbb{R}^{C \times H_s \times H_s}$ denote the query feature map, the RoI feature map and the support feature map, respectively. Note that $C$, $H_q$, $H_r$, and $H_s$ are the channel and sizes of the feature maps.

\input{figures/OSOL_framework}

\subsection{Concatenation-Based Methods}

Concatenation-based methods simply adopt the concatenation operation to aggregate $\phi_q$ and $\phi_s$, which are mainly adopted by early OSOL methods~(SiamMask~\cite{Ivan2018SiamMask}, OSCD~\cite{FuZ0S21}, FOC OSOL~\cite{Yang2021foc} and OSOLwT~\cite{Li2021OSODwF}), as shown in \autoref{fig:osol_concate_methods}.

\noindent $\bullet$ \textbf{SiamMask}~\cite{Ivan2018SiamMask}. SiamMask is one of the early deep-learning-based methods for OSOL, which concatenates $\phi_q$ with the absolute difference between $\phi_q$ and the pooled embedding vector $v_s \in \mathrm{R}^{C}$ of $\phi_s$ to generate the aggregated feature map $\phi_{\rm fused} \in \mathbb{R}^{2C \times H_q \times H_q}$, as shown in \autoref{equa:SiamMask_fuse}.
In SiamMask, $\phi_{\rm fused}$ is directly used for further components~(RPN, RoI layer) in Faster R-CNN without other modifications. SiamMask does not achieve satisfying performance since it tackles a segmentation task simultaneously. Nevertheless, as the first method for OSOL, SiamMask proposes a benchmark based on MS COCO dataset for performance comparison, pioneering many future works on OSOL and establishing a baseline for future work.

\begin{equation}
\label{equa:SiamMask_fuse}
    \phi_{\rm fused} = [\phi_q, |\phi_q - {\rm Pool}(\phi_s)|] \text{.}
\end{equation}

\noindent $\bullet$ \textbf{OSCD}~\cite{FuZ0S21}. Different from SiamMask, OSCD directly concatenates $\phi_q \in \mathbb{R}^{C \times H_q \times H_q}$ with the pooled embedding vector $v_s$ of $\phi_s \in \mathbb{R}^{C \times H_s \times H_s}$ to generate $\phi_{\rm fused} \in \mathbb{R}^{2C \times H_q \times H_q}$, as shown in \autoref{equa:OSCD_fuse}.
Besides, OSCD further conducts RoI feature aggregation to leverage the information of $\phi_s$ to facilitate the prediction of RoIs, which concatenates the RoI feature map $\phi_r$ and $\phi_s$ in depth.
OSCD proposes another OSOL benchmark based on PASCAL VOC dataset for evaluation, and it outperforms SiamFC and SiamRPN by a large margin on this benchmark.

\begin{equation}
\label{equa:OSCD_fuse}
    \phi_{\rm fused} = [\phi_q, {\rm Pool}(\phi_s)] \text{.}
\end{equation}

\noindent $\bullet$ \textbf{OSOLwT}~\cite{Li2021OSODwF} and \textbf{FOC OSOL}~\cite{Yang2021foc}. They add convolutional blocks into the concatenated features, which capture the relation between different feature units for performance improvement, as shown in \autoref{equa:FOC_OSOL_fuse}.

\begin{equation}
\label{equa:FOC_OSOL_fuse}
    \phi_{\rm fused} = {\rm Conv}([\phi_q, |\phi_q - {\rm Pool}(\phi_s)|]) \text{.}
\end{equation}


\noindent $\bigstar$ \textbf{Discussion of Concatenation-Based Methods.} Concatenation-based methods are mainly adopted by early OSOL methods. SiamMask and OSCD are the earliest concatenation-based methods for feature aggregation, while FOC OSOL and OSOLwT extend SiamMask and OSCD with convolutional blocks and some other elaborated training strategies.
However, the limitation of concatenation-based methods is that they simply aggregate features without fully excavating the relation between different local parts of two feature maps, thus impairing the matching between the foreground parts of query feature map with support feature map.

\subsection{Attention-Based Methods}

Attention-based methods take advantage of the correspondence between different parts of the support features and the query features, as shown in \autoref{fig:osod_attention_methods}.

\noindent $\bullet$ \textbf{CoAE}~\cite{HsiehLCL19}. CoAE is the first attention-based OSOL method, which proposes two operations for integral feature aggregation: co-attention~(ca) operation and co-excitation~(ce) operation. The co-attention operation is implemented using the non-local operation~\cite{wang2018non}~(an attention operation), which aggregates two feature maps according to their element-wise attention:
\begin{equation}
    \phi_{\rm fused}^{\mathrm{ca}} = \phi_q \oplus \psi(\phi_q, \phi_s) \text{,}
\end{equation}
where $\psi$ denotes the non-local operation and $\phi_{\rm fused} \in \mathbb{R}^{C \times H_q \times H_q}$. The co-excitation operation generates $\phi_{\rm fused}^{\mathrm{ce}} \in \mathbb{R}^{C \times H_q \times H_q}$ by aggregating $\phi_q$ with the pooled embedding vector $v_s \in \mathbb{R}^{C}$ of $\phi_s$ with a channel-wise multiplication:
\begin{equation}
    \phi_{\rm fused}^{\mathrm{ce}} = \phi_q \otimes {\rm Pool}(\phi_s) \text{.}
\end{equation}
CoAE adopts both these two operations for integral feature aggregation.
Besides, CoAE proposes a proposal ranking loss to supervise RPN based on RoI feature aggregation.
CoAE outperforms SiamMask on the MS COCO benchmark and OSCD on the PASCAL VOC benchmark, demonstrating the capacity of the attention mechanism on OSOL.

\input{figures/OSOL_concate_methods}

\noindent $\bullet$ \textbf{BHRL}~\cite{yang2022CVPR}, \textbf{ABA OSOL}~\cite{hsieh2023WACV}, \textbf{ADA OSOL}~\cite{zhang2022neuro}, and \textbf{AUG OSOL}~\cite{du2022neuro}. These later methods follow the co-attention and co-excitation operations in CoAE with some elaborated modifications.


\noindent $\bullet$ \textbf{AIT}~\cite{Chen2021AIT}, \textbf{CAT}~\cite{weidongCAT2021}, \textbf{SaFT}~\cite{zhao2022CVPR}. With the wide usage of transformers~\cite{Cem2021Atten} in computer vision, some methods~(AIT, CAT, SaFT) adopt multi-head attention into OSOL for feature aggregation. These methods flatten the query feature map $\phi_q$ and the support feature map $\phi_s$ to be feature sequences $\phi_q^{'} \in \mathbb{R}^{C \times H_q H_q}$ and $\phi_s^{'} \in \mathbb{R}^{C \times H_s H_s}$, then generates $\phi_{\rm fused} \in \mathbb{R}^{C \times H_q \times H_q}$ using multi-head attention to capture bidirectional correspondence between grids of them.


\noindent $\bigstar$ \textbf{Discussion of Attention-Based Methods.} Compared to attention-based methods with transformer, attention-based methods with co-attention require fewer extra parameters and less computation cost. However, CoAE is an early OSOL method, and the simple non-local operation is not enough for feature aggregation of current OSOL methods. Actually, recent methods of this type integrate co-attention with other elaborated operations to improve their performance.
On the other hand, methods based on transformer significantly improve performance, and they can easily integrate other elaborated variants of transformer structure into this framework for further performance improvement. However, current transformer-based methods bring too much extra computation cost into model training. Therefore, the efficient transformer structure is expected to be adopted for the trade-off between performance and computation cost.

\input{figures/OSOL_attention_methods}

\subsection{Transformation-Based Methods}

OS2D~\cite{Osokin2020OS2D} proposes a transformation-based method for feature aggregation, which conducts feature map transformation to match query feature map and support feature map. Given the query feature map $\phi_q$ and the support feature map $\phi_s$, OS2D first computes a 4D correlation matrix of shape $\mathbb{R}^{H_q \times H_q \times H_s \times H_s}$ which represents the correspondence between all pairs of locations from these two feature maps. Then, it uses a pre-trained TransformNet \cite{Rocco2018Trans} to generate a transformation matrix that spatially aligns the support feature map with the query feature map.
Finally, the classification score of each location of the query feature map is obtained from the combination of the correlation matrix and the transformation matrix.

\noindent $\bigstar$ \textbf{Discussion of OSOL Methods.} To sum up, concatenation-based methods are easy to implement, and they require smaller computation cost, but they have poorer performance.
Attention-based methods can capture the correspondence between support images and the foreground of query images, thus outperforming concatenation-based methods. The weakness of attention-based methods is that it is more complicated to implement them, and they require larger computation cost.
Transformation-based methods make the decision process of OSOL more interpretable, but they require a large pre-trained model to capture the spatial correspondence between query and support images.

%% file: figures/OSOL_framework.tex
\begin{figure}
\centering
\includegraphics[width=\textwidth]{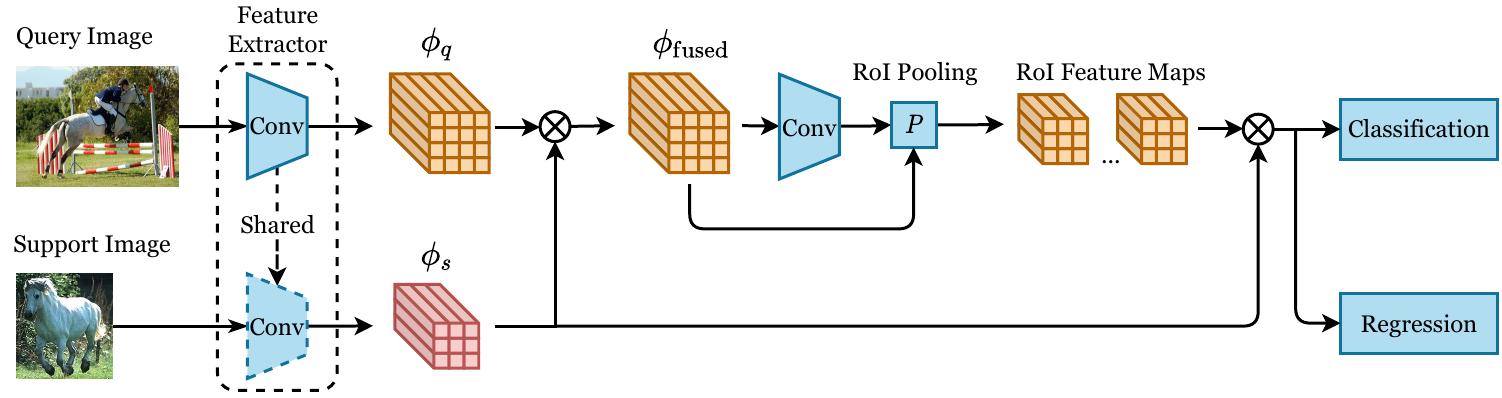}\\
\caption{The overall framework of One-Shot Object Localization~(based on Faster R-CNN). The model takes a query image and a support image as inputs, then uses a siamese convolutional feature extractor to extract the query feature map and the support feature map. Then it applies \textit{integral feature aggregation} to aggregate these two feature maps into a fused feature map and forwards it into RPN and RoI layer to generate region proposals and RoI features, respectively. The aggregation method is implemented differently in different OSOL methods. Finally, the RoI features are used for the classification task and the regression task. Some methods additionally apply \textit{RoI feature aggregation} to aggregate the RoI features with the support features.}
\label{fig:osol_framework}
\end{figure}

%% file: figures/OSOL_concate_methods.tex
\begin{figure}
\centering
\subfigure[SiamMask~\cite{Ivan2018SiamMask}]{
    \begin{minipage}[t]{0.45\linewidth}
    \centering
    \includegraphics[width=6.5cm]{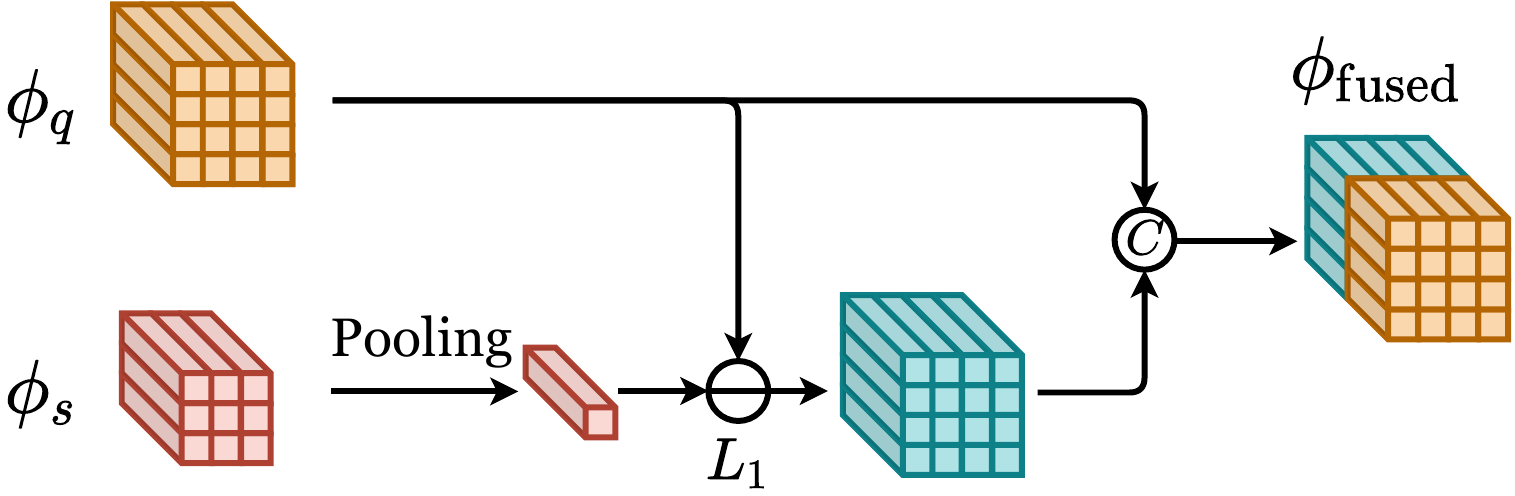}
    \end{minipage}%
}%
\subfigure[OSCD~\cite{FuZ0S21} OSOLwT~\cite{Li2021OSODwF}]{
    \begin{minipage}[t]{0.55\linewidth}
    \centering
    \includegraphics[width=7cm]{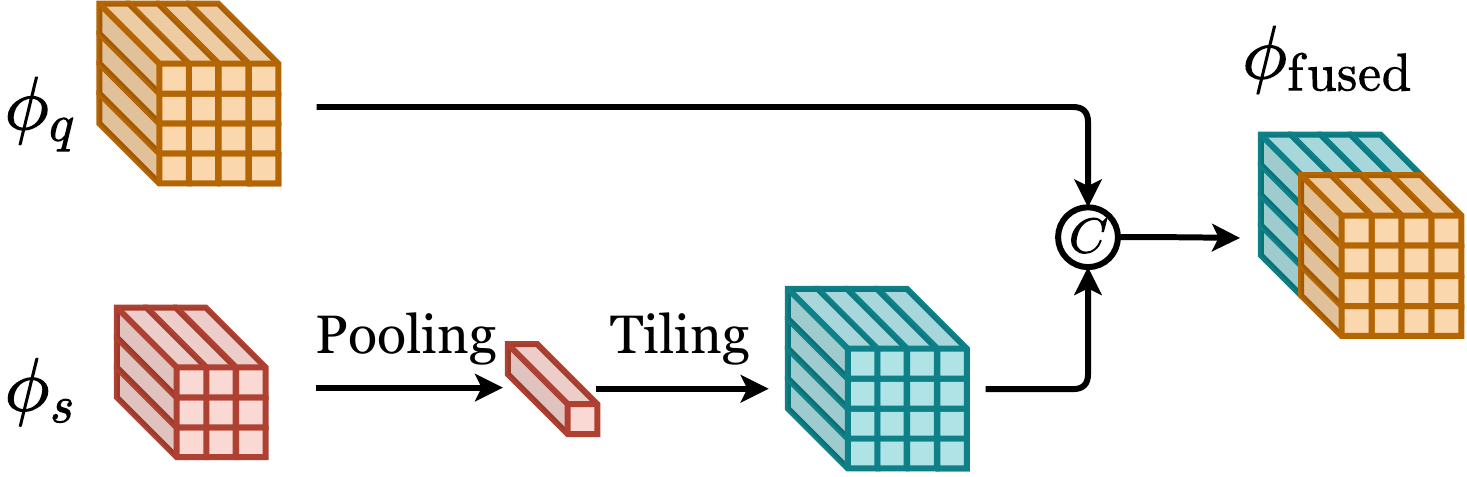}
    \end{minipage}%
}%

\subfigure[FOC OSOL~\cite{Yang2021foc}]{
    \begin{minipage}[t]{\linewidth}
    \centering
    \includegraphics[width=8cm]{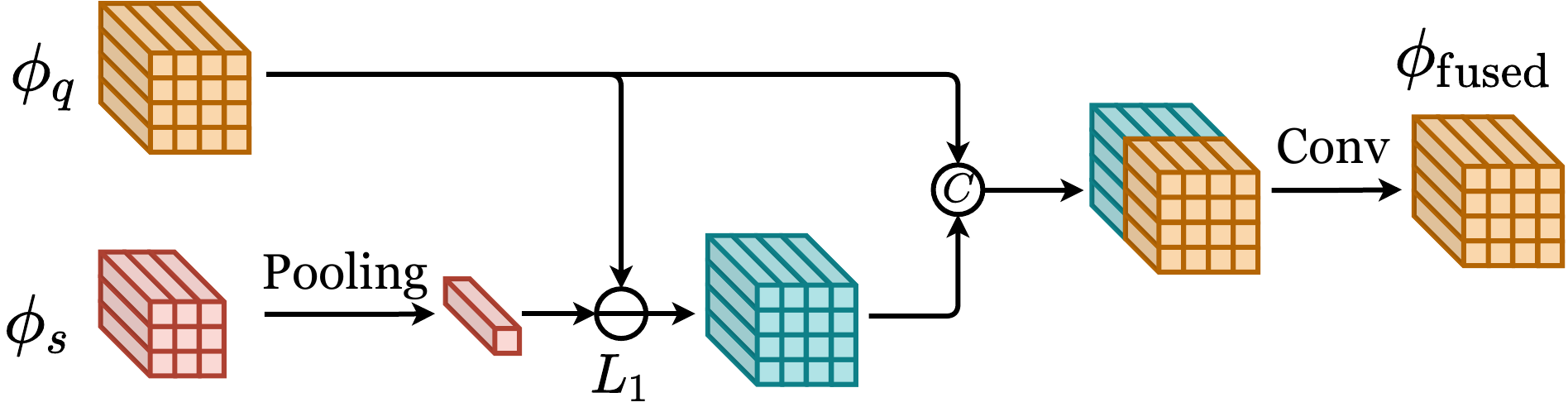}
    \end{minipage}%
}
\centering
\caption{Overview of concatenation-based integral feature aggregation methods in OSOL. OSCD and OSOLwT concatenate the query feature map with the pooled embedding vector of the support feature map. SiamMask concatenates the query feature map with the absolute difference between the query feature map and the pooled embedding vector of the support feature map instead. FOC OSOL additionally applies convolution blocks on the integral feature map generated in SiamMask.}
\label{fig:osol_concate_methods}
\end{figure}

%% file: figures/OSOL_attention_methods.tex
\begin{figure}
\centering
\subfigure[CoAE, BHRL, ABA OSOL, ADA OSOL, AUG OSOL]{
    \begin{minipage}[t]{0.4\linewidth}
    \centering
    \includegraphics[width=6cm]{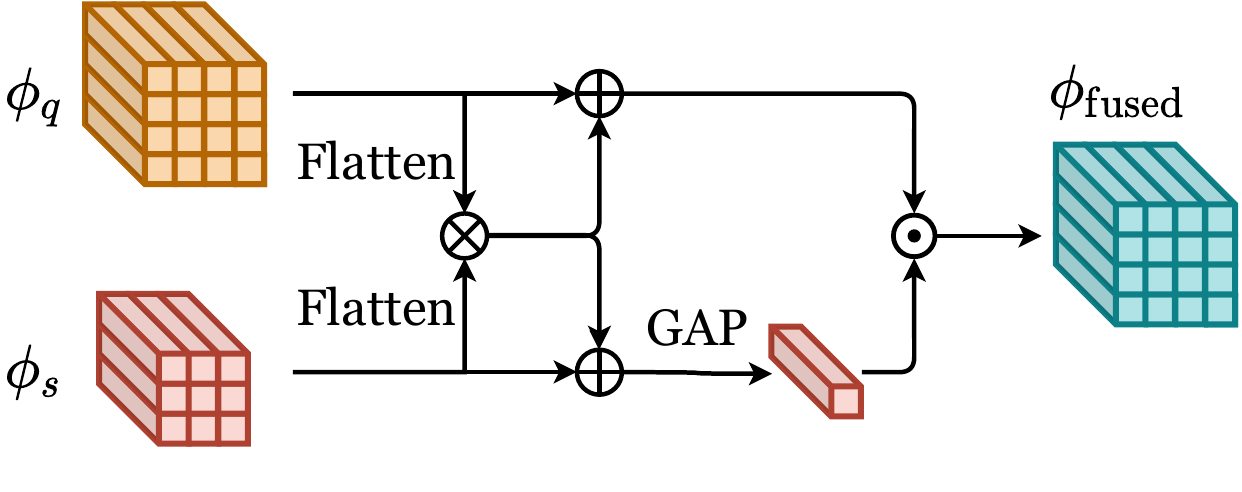}
    \end{minipage}%
}%
\subfigure[AIT, CAT, SaFT]{
    \begin{minipage}[t]{0.6\linewidth}
    \centering
    \includegraphics[width=7.5cm]{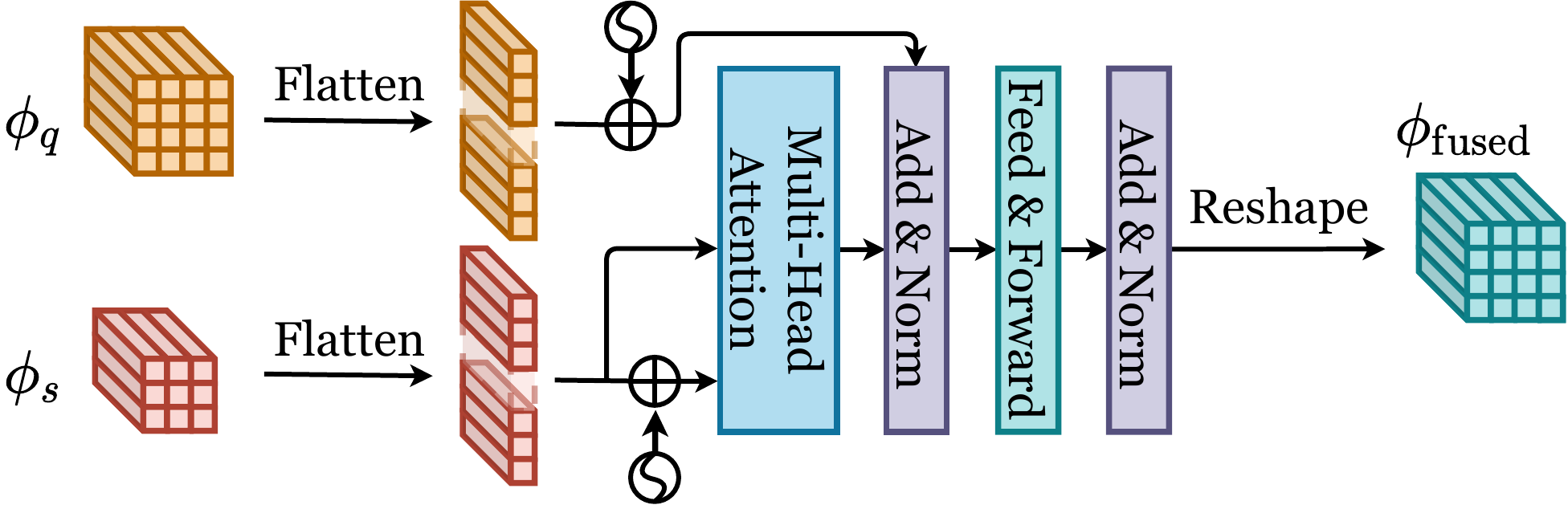}
    \end{minipage}%
}%
\centering
\caption{Overview of attention-based integral feature aggregation methods in OSOL. Some methods~(CoAE, BHRL, ABA OSOL, ADA OSOL, AUG OSOL) use a non-local operation for integral feature aggregation, while some methods~(AIT, CAT, SaFT) use transformer to capture attention between query and support images.}
\label{fig:osod_attention_methods}
\end{figure}

%% file: n03_FSOD.tex
\section{Standard Few-Shot Object Detection\label{sec:standard_fsod}}

\textbf{Task Settings.} The previous OSOL setting guarantees that every query image contains objects with the same category as the given support image, i.e., the model knows precisely the object classes contained in each test image.
However, this setting is not realistic in the real world, and a more challenging LSOD setting, named Few-Shot Object Detection~(FSOD), is adopted by more papers.
This section first introduces the standard FSOD, and other FSOD settings~(named extensional FSOD) are based on the standard FSOD, which will be analyzed in the later sections.
Specifically, the base dataset~($D_B$) of standard FSOD consists of abundant annotated instances of base classes $C_B$, and the novel dataset~($D_N$) consists of scarce annotated instances of novel classes $C_N$~($C_B$ and $C_N$ are not intersected). During testing, the model is evaluated on the test dataset comprising objects of both base classes and novel classes. The differences between FSOD and OSOL are as below:

\begin{enumerate}
    \item Since OSOL precisely knows the object categories contained in each test image, it only requires a binary classification task to discriminate whether the potential object is foreground or background according to the given support image. In contrast, FSOD requires a multi-class classification task to predict the category of the potential object.
    \item OSOL samples support images independently for each test image, FSOD samples the support images only once for all test images.
    \item The shot number of support images per category can be larger than one in FSOD.
\end{enumerate}


\noindent \textbf{Method Categorization.} Current standard FSOD methods can be categorized into fine-tune-based methods and fine-tune-free methods.
Most methods are fine-tune-based methods, which require to fine-tune the model on the novel dataset for the significant improvement of performance.
Early fine-tune-based methods adopt the scheme of meta-learning, and they also concentrate on the methods of feature aggregation as OSOL methods. Furthermore, the increased number of annotated samples opens up the possibility for standard FSOD methods to adopt the scheme of transfer-learning, which pre-trains an object detector on the base dataset and fine-tunes this pre-trained model for novel classes on the novel dataset.
Early transfer-learning methods like LSTD~\cite{chen2018lstd} are outperformed by the meta-learning methods in that period until the emergence of TFA \cite{wang2020few}.
On the other hand, fine-tune-free methods aim to remove the fine-tuning step because fine-tuning step is not suitable for FSOD in real life for its nonnegligible computation cost.
In this survey, the meta-learning methods are first analyzed since they are highly correlated to OSOL methods, then transfer-learning methods and fine-tune-free methods are analyzed later.

\subsection{Meta-Learning Methods}

Similar to OSOL, the meta-learning methods for standard FSOD first define a large number of few-shot detection tasks on the base dataset to train the model. The difference is that each few-shot task contains a query image and multiple support images since FSOD requires support images from all base classes for the multi-class classification task.
Another difference is that meta-learning methods for standard FSOD have an additional fine-tuning stage that OSOL methods lack, which continues to meta-train the model by sampling support images from both base classes and novel classes for each few-shot task.
The meta-learning framework of standard FSOD is similar to that of OSOL, which conducts ``integral feature aggregation'' and ``RoI feature aggregation'' to aggregate the query features with support features to incorporate the information of support images into the query image for prediction.
Early meta-learning methods only conduct RoI feature aggregation, and later methods conduct both integral and RoI feature aggregation~(named ``mixed feature aggregation'') for better performance.
Therefore, meta-learning methods for standard FSOD are categorized into RoI feature aggregation methods and mixed feature aggregation methods for a more explicit presentation in this survey.

\subsubsection{\textbf{RoI Feature Aggregation Methods}}

RoI feature aggregation methods aggregate the RoI features with support features to generate class-specific RoI features for the classification and regression tasks. Unlike OSOL methods that almost all adopt Faster R-CNN as the detection framework, early meta-learning methods explore RoI feature aggregation methods on both single-stage and two-stage detectors.
These RoI feature aggregation methods can be categorized into two types according to the type of aggregated features: \textbf{RoI feature-vector aggregation methods}~(FSRW~\cite{kang2019few}, Meta R-CNN~\cite{yan2019meta}, CME~\cite{Li_2021_CVPR}, TIP~\cite{Li2021TIP}, VFA~\cite{han2023few}, FSOD-KT~\cite{geonukkimKnowledge2020}, GenDet~\cite{liu2022tnnls}, FsDet~\cite{xiao2020few}, DRL~\cite{liu2021dynamic}, and AFD-Net~\cite{liu2020afd}) and \textbf{RoI feature-map aggregation methods}~(Attention-RPN~\cite{Fan0TT20}, QA-FewDet~\cite{Han2021ICCV}, KFSOD~\cite{zhang2022CVPR}, PNSD~\cite{zhang2020few}, MM-FSOD~\cite{han2022multimodal}, SQMG-FSOD~\cite{Zhang2021SQMG}, ICPE~\cite{lu2022breaking}, DAnA-FasterRCNN~\cite{tungchenHeadTail2021}, TENET~\cite{zhang2022ECCV}, Hierarchy-FasterRCNN~\cite{park2022hierarchical}, IQ-SAM~\cite{lee2022WACV}, and Meta Faster R-CNN~\cite{guangxingMetaFaster2021}).

\input{figures/fsod_meta_roi_aggregation}

The ``RoI feature-vector aggregation methods'' can be categorized into two types, which are first proposed by \textbf{FSRW} and \textbf{FsDet}, respectively.

\noindent $\bullet$ \textbf{FSRW}~\cite{kang2019few} is the first meta-learning method for standard FSOD based on the YOLOv2 detection framework. FSRW simply aggregates each feature vector $v_r \in \mathbb{R}^{C}$ at each pixel of the query feature map with the pooled embedding $v_s \in \mathbb{R}^{C}$ of the support feature map, aiming to highlight the important features corresponding to the support image using a simple element-wise multiplication:
\begin{equation}
\label{equa:fsod_meta_element_multiplication}
v_{\rm fused} = v_r \otimes v_s \text{.}
\end{equation}
The fused feature vector $v_{\rm fused} \in \mathbb{R}^{C}$ is used to predict the classification score~(for the class that $v_s$ is from) and location regression, as shown in \autoref{fig:fsod_meta_roi_aggregation_fsrw}.
\textbf{Meta R-CNN}~\cite{yan2019meta}, \textbf{CME}~\cite{Li_2021_CVPR}, \textbf{TIP}~\cite{Li2021TIP}, \textbf{VFA}~\cite{han2023few}, \textbf{FSOD-KT}~\cite{geonukkimKnowledge2020} and \textbf{GenDet}~\cite{liu2022tnnls} follow this simple element-wise multiplication operation with other elaborated extensions.

\noindent $\bullet$ \textbf{FsDet}~\cite{xiao2020few} upgrades this simple element-wise multiplication operation to a more complex yet effective version, as shown in \autoref{fig:fsod_meta_roi_aggregation_fsdet}. Specifically, given the RoI feature vector $v_r$ and the support feature vector $v_s$, the aggregated feature vector $v_{\rm fused}$ is calculated as the concatenation of their linearly transformed element-wise multiplication, subtraction and the original $v_r$, as shown in \autoref{equa:fsod_meta_multiform_aggregation}~(${\rm FC}$ denotes a fully-connected layer that reduces the dimension).
With this extended aggregation method, FsDet outperforms Meta R-CNN on both MS COCO benchmark and PASCAL VOC benchmark.

 \begin{equation}
\label{equa:fsod_meta_multiform_aggregation}
    v_{\rm fused} = [{\rm FC}(v_r \otimes v_s), {\rm FC}(v_r - v_s), v_r] \text{,}
\end{equation}

\noindent $\bullet$ \textbf{AFD-Net}~\cite{liu2020afd} and \textbf{DRL}~\cite{liu2021dynamic}. These two methods follow FsDet in this RoI feature-vector aggregation method with some other modifications.



Unlike the above \textbf{RoI feature-vector aggregation methods} which concentrate on the aggregation of feature vectors, \textbf{RoI feature-map aggregation methods} focus on the aggregation of feature maps that preserves spatial information for better excavating the relation between query and support images. Some methods only adopt simple \textbf{concatenation operation and element-wise operation} for the feature map aggregation, while newly proposed methods tend to adopt \textbf{attention operation} for feature map aggregation.

\noindent $\bullet$ \textbf{Concatenation operation \& element-wise operation for RoI feature-map aggregation.}
SQMG-FSOD~\cite{Zhang2021SQMG} simply concatenates the RoI feature map with the support feature map for the RoI feature-map aggregation. While some methods~(Attention-RPN~\cite{Fan0TT20}, QA-FewDet~\cite{Han2021ICCV}, KFSOD~\cite{zhang2022CVPR}, PNSD~\cite{zhang2020few}, FCT~\cite{han2022CVPR}, and MM-FSOD~\cite{han2022multimodal}) utilize a multi-relation head that adopt both concatenation operation and element-wise operation. Specifically, this multi-relation head consists of a global-relation head, a patch-relation head, and a local-relation head. The global-relation head concatenates $\phi_r$ and $\phi_s$ in depth with a pooling operation. The patch-relation head concatenates $\phi_r$ and $\phi_s$ with several convolutional blocks on it. And the local-relation head aggregates $\phi_r$ and $\phi_s$ by calculating the pixel-wise and depth-wise similarities between them.
These methods conduct both integral and RoI feature aggregation, which will be specified later.

\noindent $\bullet$ \textbf{Attention operation for RoI feature-map aggregation.}
Some methods~(ICPE~\cite{lu2022breaking}, DAnA-FasterRCNN~\cite{tungchenHeadTail2021}, TENET~\cite{zhang2022ECCV}, Hierarchy-FasterRCNN~\cite{park2022hierarchical}, IQ-SAM~\cite{lee2022WACV}, and Meta Faster R-CNN~\cite{guangxingMetaFaster2021}) adopt the attention operation to conduct RoI feature-map aggregation. Specifically, they calculate the aggregated feature map according to the similarity score~(attention) between each pair of elements from $\phi_r$ and $\phi_s$.
In these methods, ICPE conducts only RoI feature aggregation with some proposed modifications. Specifically, it additionally incorporates the information of query images into support images before the final feature aggregation, and it adjusts the importance of different support images instead of treating them as equals.
Other methods conduct both integral and RoI feature aggregation, which will be specified later.



\noindent $\bigstar$ \textbf{Discussion of RoI Feature Aggregation Methods.} RoI Feature Aggregation Methods are categorized into RoI feature-vector aggregation methods and RoI feature-map aggregation methods. RoI feature-vector aggregation methods are early meta-learning methods for FSOD, whose approaches are simple and limit their performance. On the other hand, RoI feature-map aggregation methods preserve spatial information of query and support samples, towards fully extracting the spatial relations between query and support features. Therefore, RoI feature-map aggregation methods can better discriminate features of different objects and achieve higher performance.

\subsubsection{\textbf{Mixed Feature Aggregation Methods}}

The above section discusses only the RoI feature aggregation, while most newly proposed methods~(named ``mixed feature aggregation methods'') additionally conduct integral feature aggregation to incorporate class-specific information into the shallow components of the detection model.
The integral feature aggregation methods are mainly conducted on the feature-maps~(not feature-vectors) and can be categorized into \textbf{concatenation \& element-wise operations}~(Attention-RPN~\cite{Fan0TT20}, QA-FewDet~\cite{Han2021ICCV}, KFSOD~\cite{zhang2022CVPR}, PNSD~\cite{zhang2020few}, MM-FSOD~\cite{han2022multimodal}, Meta Faster R-CNN~\cite{guangxingMetaFaster2021}), \textbf{convolutional operation}~(SQMG-FSOD~\cite{Zhang2021SQMG}), and \textbf{attention operation}~(DAnA-FasterRCNN~\cite{tungchenHeadTail2021}, TENET~\cite{zhang2022ECCV}, Hierarchy-FasterRCNN~\cite{park2022hierarchical}, IQ-SAM~\cite{lee2022WACV}, DCNet~\cite{hu2021dense}, Meta-DETR~\cite{gongjiezhangMetaDETR2021}, FCT~\cite{han2022CVPR}).

\noindent $\bullet$ \textbf{Concatenation \& element-wise operations for integral feature aggregation.} Attention-RPN~\cite{Fan0TT20} conducts integral feature map aggregation by using $\phi_s \in \mathbb{R}^{C \times H_s \times H_s}$ as a kernel and sliding it across $\phi_q \in \mathbb{R}^{C \times H_q \times H_q}$ to compute similarities at each location. Specifically, the element at the location $(c, h, w)$ of the aggregated feature map $\phi_{\rm fused}$ is calculated in \autoref{equa:fsod_meta_integral_element_multiplication}~(note that $i,j \in \{1, \cdots, H_s \}$).
Some methods~(QA-FewDet~\cite{Han2021ICCV}, KFSOD~\cite{zhang2022CVPR}, PNSD~\cite{zhang2020few}, MM-FSOD~\cite{han2022multimodal}, Meta Faster R-CNN~\cite{guangxingMetaFaster2021}) follow this integral feature aggregation method with other extensions.

\begin{equation}
\label{equa:fsod_meta_integral_element_multiplication}
    {\phi_{\rm fused}}_{(c,h,w)} = \sum\limits_{i,j} {\phi_q}_{(c,h+i-1,w+j-1)} \cdot {\phi_s}_{(c,i,j)} \text{.}
\end{equation}

\noindent $\bullet$ \textbf{Convolutional operation for integral feature aggregation.} SQMG-FSOD~\cite{Zhang2021SQMG} proposes another integral feature aggregation method by generating convolutional kernels from support features and using the generated kernels to enhance query features.
Furthermore, SQMG-FSOD not only learns a distance metric to compare RoI features and support features for filtering out irrelevant RoIs but also utilizes this metric to assign weights to support samples by comparing them with query images. 
Additionally, it proposes a hybrid loss to mitigate the false positive problem~(i.e., some background RoIs are misclassified into objects).

\noindent $\bullet$ \textbf{Attention operation for integral feature aggregation.} Newly proposed methods~(DCNet~\cite{hu2021dense}, DAnA-FasterRCNN~\cite{tungchenHeadTail2021}, TENET~\cite{zhang2022ECCV}, Hierarchy-FasterRCNN~\cite{park2022hierarchical}, IQ-SAM~\cite{lee2022WACV}, Meta-DETR~\cite{gongjiezhangMetaDETR2021}, and FCT~\cite{han2022CVPR}) tend to adopt attention operation for integral feature aggregation. Attention operation aggregates two feature-maps using a similar manner as scaled dot-product attention~\cite{Cem2021Atten}. It extracts the key map and the value map from the query image and the support image, respectively, then calculates the pixel-wise similarities between these two key maps and uses them to aggregate two value maps.

\begin{itemize}
    \item Meta-DETR also adopts attention operation for integral feature aggregation with a significant boost in performance. The major difference is that it adopts Deformable DETR~\cite{zhu2021deformable} as the detection framework. DETR is an end-to-end transformer-based detector that eliminates anchor boxes in former detectors. Besides, Meta-DETR proposes a correlational aggregation module~(CAM) that uses single-head attention to aggregate the query feature-maps with the support feature-maps. The aggregated features are finally fed into a class-agnostic transformer to predict object categories and locations.
    
    \item Most of these methods aggregate the query and support features that are extracted from the backbone independently, while FCT surpasses this limit and instead aggregates the features in each layer of the ViT backbones, which achieves significant performance improvement.
    First, it splits query images and support images into image tokens and add position \& branch embeddings into them~(i.e., position embedding discriminates the position of the token, and branch embedding discriminates whether the token is from support image or query image).
    Next, it concatenates all query and support tokens into a sequence and feeds them into a transformer to generate the aggregated integral features.
\end{itemize}

\noindent $\bigstar$ \textbf{Discussion of Mixed Feature Aggregation Methods.} Compared to RoI feature aggregation methods, mixed feature aggregation methods additionally conduct integral feature aggregation to incorporate category-specific information into the shallow components~(mainly RPN) of the detection model, which extracts more positive region proposals for the further classification \& regression tasks and improves the performance.
Mixed feature aggregation methods are categorized into three types: \textbf{concatenation \& element-wise operations}, \textbf{convolutional operation}, and \textbf{attention operation}.
Simple concatenation \& element-wise operations are mostly adopted by early FSOD methods, which have poor performance and need to combine other components altogether for performance improvement.
Convolutional operation is still simple, which cannot fully incorporate the information of support features into query features.
Attention operation captures the relation between local regions in query feature maps and support feature maps, which better discriminates different local regions, and these methods overall achieve better performance.



\subsubsection{\textbf{Other Meta-Learning Methods}}

There are some other meta-learning methods that focus on issues other than the aggregation method of features, which are \textbf{weight-prediction-based methods} and \textbf{metric-learning-based methods}.

\noindent $\bullet$ \textbf{Weight-Prediction-Based Methods.} MetaDet~\cite{wangyxMetaTo2019} proposes a meta-learning method that learns to predict the weights of category-specific components of the model. MetaDet predicts category-specific~(e.g., the classification and regression branches) weights for novel classes from few samples and fine-tunes the model on the novel dataset.
Meta-RetinaNet~\cite{Li2020MetaRetina} is another method which adopts RetinaNet as the detection framework and predicts the weights of the whole network.


\noindent $\bullet$ \textbf{Metric-Learning-Based Methods.} IR-FSOD~\cite{huang2021instant} directly learns to compare the similarity between the RoI features with support features from different classes to generate the classification scores.
CAReD~\cite{quan2022displays} also adds another metric learning branch for classification apart from the main classification branch.


\subsection{Transfer-Learning Methods}

Transfer-learning methods regard FSOD as a transfer-learning problem in which the source domain is the base dataset, and the target domain is the novel dataset. Current transfer-learning methods mainly adopt Faster R-CNN as the detection framework, consisting of two stages: base training and few-shot fine-tuning, as shown in \autoref{fig:transfer_learning_fsod}. The base training stage trains an object detector on the base dataset. After this stage, the object detector will obtain an effective feature extractor and achieve good performance on base classes. Then, in the few-shot fine-tuning stage, this pre-trained object detector will be fine-tuned on the novel dataset to detect novel classes. In this way, the common knowledge for feature extraction and proposal generation can be transferred from base classes to novel classes.

\input{figures/fsod_transfer_learning}

\noindent $\bullet$ \textbf{LSTD}~\cite{chen2018lstd} is the first method to adopt the transfer-learning scheme for FSOD. It adopts Faster R-CNN as the detection framework with two regularization terms in the few-shot fine-tuning stage.
Specifically, the first term suppresses background regions in the feature maps, and the second term promotes the fine-tuned model to generate similar predictions with the source model.
Regrettably, the performance of LSTD is exceeded by the meta-learning methods during the same period.

\noindent $\bullet$ \textbf{TFA}~\cite{wang2020few}~(Two-Stage Fine-tuning Approach) significantly improves the performance of transfer-learning methods based on the Faster R-CNN detection framework. In the base training stage, TFA pre-trains the model on the base dataset as previous transfer-learning methods. Then, in the few-shot fine-tuning stage, it freezes the main components of Faster R-CNN and only fine-tunes the last two layers (box classification and regression layers) of Faster R-CNN. The loss function used in the few-shot fine-tuning stage is the same as the base training stage but with a lower learning rate. The dataset used in the few-shot fine-tuning stage is a balanced dataset containing a few training samples of novel classes and a few selected training samples of base classes. This design retains the model's detection ability for base classes and mitigates the problem that some objects of base classes are misclassified into novel classes. With this simple but effective training strategy, TFA outperforms early meta-learning methods like FSRW, MetaDet, and Meta R-CNN on both MS COCO benchmark and PASCAL VOC benchmark.

\noindent $\bullet$ DeFRCN~\cite{DeFRCN2021Qiao} significantly improves the performance of TFA with two concise modifications: (1) DeFRCN assigns different importance values to the gradients from RPN module and R-CNN module, which is motivated by the viewpoint that RPN module and R-CNN module may learn paradoxically and the learning of these two modules should be decoupled. (2) DeFRCN utilizes a pre-trained classifier as an auxiliary branch for the classification of region proposals.
DeFRCN further validates the effectiveness of transfer-learning methods for FSOD, and many methods are proposed following this transfer-learning paradigm. In this survey, transfer-learning methods are categorized into \textbf{feature-augmentation-based methods}, \textbf{classification-based methods}, \textbf{regression-based methods}, \textbf{RPN-based methods}, \textbf{data-augmentation-based methods}, and \textbf{pre-train-based methods} according to the detection stage they focus on.


\subsubsection{\textbf{Feature-Augmentation-Based Methods}}

Feature-augmentation-based methods focus on the feature extraction stage of an FSOD model. These methods apply different augmentations to the features, aiming to better transfer the features learned on the base dataset to the novel dataset. Current feature-augmentation-based methods can be categorized into three types: \textbf{self-attention-based methods}~(CT-FSOD~\cite{yang2020context}, AttFDNet~\cite{chen2020leveraging}), \textbf{feature-discretization-based methods}~(SVD-FSOD~\cite{wu2021NIPS}, KD-FSOD~\cite{pei2022ECCV}), and \textbf{feature-inheritance-based methods}~(${\rm FSOD}^{\rm up}$~\cite{wu2021universal}, FADI~\cite{cao2021NIPS}).

\noindent $\bullet$ \textbf{Self-Attention-Based Methods for Feature-Augmentation.}
Self-attention-based methods~(CT-FSOD~\cite{yang2020context}, AttFDNet~\cite{chen2020leveraging}) adopt self-attention to augment the extracted features.

\noindent $\bullet$ \textbf{Feature-Discretization-Based Methods for Feature-Augmentation.}
Feature-discretization-based methods~(SVD-FSOD~\cite{wu2021NIPS}, KD-FSOD~\cite{pei2022ECCV}) discretize the feature-map by projecting each pixel of the feature-map into a learned codebook~(i.e., replacing each pixel of the feature-map with its nearest code), thus enhancing the discrimination of features from different categories.



\noindent $\bullet$ \textbf{Feature-Inheritance-Based Methods for Feature-Augmentation.}
Feature-inheritance-based methods~(${\rm FSOD}^{\rm up}$~\cite{wu2021universal}, FADI~\cite{cao2021NIPS}) inherit the features of base classes to the features of novel classes for augmentation, which mitigates the data scarcity problem of novel classes.

\noindent $\bigstar$ \textbf{Discussion of Feature-Augmentation-Based Methods.} Self-attention-based methods incorporate interpretability into the decision-making of FSOD through the attention heatmaps.
However, self-attention-based methods are early FSOD methods, and the attention operations they adopt are primitive, restricting their performance.

Feature-discretization-based methods utilize feature discretization to enhance the discrimination of features from different categories, but they haven't demonstrated the visual concepts that the discretized features represent. 
Besides, KD-FSOD requires an additional step to train an extra visual-word model and needs knowledge distillation to inherit the knowledge of this visual-word model into the few-shot detector, bringing a non-negligible burden into model training.

Feature-inheritance-based methods utilize the knowledge from base classes as ``free lunch'' to augment the features of novel classes with negligible cost. However, in the scenario that base classes and novel classes are not in the same domain, it is unclear whether these methods still work since base classes and novel classes share less common knowledge.

\subsubsection{\textbf{Classification-Based Methods}}

Classification-based methods aim to improve the classification branch of the detection model. Early classification-based methods focus on improving the main classification branch with some elaborated \textbf{metric learning methods}~(RepMet~\cite{karlinsky2019repmet}, NP-RepMet~\cite{YangWSL20}, PNPDet~\cite{zhang2021pnpdet}, FSOD-KI~\cite{yang2023efficient}). New classification-based methods mostly propose another classification branch to assist the main classification branch, including \textbf{additional-classifier-based methods}~(FSCN~\cite{Li2021CRaDR}), \textbf{contrastive-learning-based methods}~(FSCE~\cite{sun2021fsce}, FSRC~\cite{shangguan2022few}, CoCo-RCNN~\cite{ma2022ECCV}), \textbf{knowledge-graph-based methods}~(KR-FSOD~\cite{wang2022few}), and \textbf{semantic-infor-mation-based methods}~(SRR-FSOD~\cite{zhu2021semantic}).

\noindent $\bullet$ \textbf{Metric Learning Methods for Classification.} These methods~(RepMet~\cite{karlinsky2019repmet}, NP-RepMet~\cite{YangWSL20}, PNPDet~\cite{zhang2021pnpdet}, FSOD-KI~\cite{yang2023efficient}) propose elaborated metric learning methods to directly improve the main classification branch. 

\noindent $\bullet$ \textbf{Additional-Classifier-Based Methods for Classification.}
FSCN~\cite{Li2021CRaDR} proposes a few-shot correction network~(FSCN) as an additional classification branch of the model, which makes class predictions for the cropped region proposals with a pre-trained image classifier. These classification scores are used to refine the classification scores from the main branch.
Besides, this paper proposes a semi-supervised distractor utilization method to select unlabeled distractor proposals for novel classes and a confidence-guided dataset pruning~(CGDP) method for filtering out training images containing unlabeled objects of novel-classes.

\noindent $\bullet$ \textbf{Contrastive-Learning-Based Methods for Classification.} These methods~(FSCE~\cite{sun2021fsce}, CoCo-RCNN~\cite{ma2022ECCV}, FSRC~\cite{shangguan2022few}) adopt contrastive learning to assist the classification of region proposals.

\begin{itemize}
    \item FSCE introduces a contrastive loss to improve the classification performance of the model.
    FSCE proposes a contrastive loss function to maximize the similarity between objects of the same category and promote the distinctiveness of region proposals from different categories. This work is the first attempt to adopt contrastive learning into transfer-learning-based FSOD, which significantly improves the performance of the baseline TFA.
    
\end{itemize}


\noindent $\bullet$ \textbf{Knowledge-Graph-Based Methods for Classification.}
KR-FSOD~\cite{wang2022few} proposes an additional classification branch based on an external knowledge graph with potential objects as nodes. The model predicts the category of each potential object according to the information of its nearby objects, which is extracted from this external knowledge graph. KR-FSOD improves the performance by incorporating the external knowledge graph into the FSOD model.

\noindent $\bullet$ \textbf{Semantic-Information-Based Methods for Classification.}
SRR-FSOD~\cite{zhu2021semantic} proposes an additional classification branch utilizing class semantic information to promote the classification, which utilizes the external semantic information into the FSOD model for higher performance.
Specifically, SRR-FSOD projects the visual features into the semantic space using a linear projection. In this semantic space, multiple word embeddings are used as semantic embeddings to represent all base and novel classes. It generates class probabilities for the projected semantic embeddings by calculating the similarities between the projected visual features and the class semantic embeddings.

\noindent $\bigstar$ \textbf{Discussion of Classification-Based Methods.} Metric learning methods are early methods for FSOD with insufficient performance compared with the latest FSOD methods, indicating that simple modification on the RoI classifier is not enough for FSOD.

Additional-classifier-based method~(FSCN) achieves a large performance improvement. However, it requires a pre-trained image classifier, resulting in an unfair comparison with other FSOD methods.

Contrastive-learning-based methods incur minimal additional cost during model training while yielding a substantial improvement in performance. Besides, they can be seamlessly integrated into other FSOD methods.

Knowledge-graph-based method~(KR-FSOD) is well motivated, but the performance is currently not promising. Additionally, like FSCN, it cannot be readily applied to novel classes in real-world FSOD applications due to the unavailability of corresponding knowledge graphs.

Semantic-information-based method~(SRR-FSOD) serves as a bridge between FSOD and zero-shot learning by incorporating class semantic information into the model.
This approach has the potential for enhancing performance with the large-scale cross-modal models.
Nevertheless, it may not be suitable for novel classes that haven't been learned before.

\subsubsection{\textbf{Regression-Based Methods}}

Regression-based methods focus on improving the regression branch of detection model.
SRR-FSD~\cite{kim2022ICPR} proposes a refinement approach to improve the regression of region proposals in RPN.
Specifically, SRR-FSD expands the regression branch into multiple successive regression heads. Each regression head receives the region proposals generated from the preceding regression head and continues to refine these region proposals for generating more positive samples.

\noindent $\bigstar$ \textbf{Discussion of Regression-Based Methods.} While the performance of the current regression-based method (SRR-FSD) is currently not ideal, it's important to note that such methods are still rare, and there is ample opportunity for future exploration and improvement.

\subsubsection{\textbf{RPN-Based Methods}}

CoRPN~\cite{zhang2020cooperating} improves the RPN in Faster R-CNN for standard FSOD. CoRPN assumes that the RPN pre-trained on base classes will miss some objects of novel classes. Therefore, it uses multiple foreground-background classifiers in RPN instead of the original single one to mitigate this problem. During testing, a given proposal box is assigned with the score from the most certain RPN. During training, only the most certain RPN will get the gradient from the corresponding bounding box. CoRPN proposes a diversity loss to encourage the diversity of these RPNs and a cooperation loss to mitigate firm rejection of foreground proposals.

\noindent $\bigstar$ \textbf{Discussion of RPN-Based Methods.} RPN-based method~(CoRPN) directly devises multiple RPNs to retrieve those missed novel objects, which addresses the problem that novel objects tend to be missed by the RPN trained on the base dataset. However, it is limited in R-CNN-based model, and it is unclear whether it still works when integrated into other FSOD models.

\subsubsection{\textbf{Data-Augmentation-Based Methods}}

Data-augmentation-methods aim to generate more samples for each novel class, thus directly tackling the data-scarce problem of few-shot setting. Current data augmentation methods can be divided into two categories: \textbf{sample generation in the input-pixel space} and \textbf{sample generation in the feature space}. The former type directly generates samples in the input-pixel space that are understandable and perceivable by humans, which can be further divided into \textbf{multi-scale augmentation methods} and \textbf{novel-instance-mining methods}. The latter type synthesizes more deep features for the novel classes, which can be further divided into \textbf{distribution inheritance methods} and \textbf{generator-based methods}.

\noindent $\bullet$ \textbf{Sample Generation In the Input-Pixel Space $\rightarrow$ Multi-Scale Augmentation Methods.}
MPSR~\cite{wu2020multi} and FSSP~\cite{xu2021few} both apply data augmentation to enrich the scales of positive samples.

\begin{itemize}
    \item MPSR claims that although feature pyramid network (FPN)~\cite{lin2017CVPR} may mitigate the scale variation issue, it cannot address the sparsity of scale distribution in FSOD.
    Therefore, MPSR proposes a strategy to directly augment the scales of objects in the input pixel space, which extracts each positive object independently and resizes them to multiple scales. The augmented multi-scale samples are fed into the RPN module and detection heads for training.
\end{itemize}

\noindent $\bullet$ \textbf{Sample Generation In the Input-Pixel Space $\rightarrow$ Novel-Instance-Mining Methods.}
MINI~\cite{cao2022mini}, PSEUDO~\cite{kaul2022CVPR}, Decoupling~\cite{gao2022decoupling}, and N-PME~\cite{liu2022ICASSP} excavate the unlabeled novel objects in the dataset for data augmentation.


\noindent $\bullet$ \textbf{Sample Generation In the Feature Space $\rightarrow$ Distribution Inheritance Methods.}
FSOD-KD~\cite{zhao2022acmmm}, PDC~\cite{li2022proposal}, and FSOD-DIS~\cite{wu2022ECCV} generate more novel features by transferring the feature distribution from the base dataset for data augmentation, which stem from the same few-shot learning method~\cite{yang2021ICLR}.
Specifically, these methods assume that the feature distribution of a class can be approximated as a Gaussian distribution and similar classes have similar feature distributions.
Therefore, they calculate the feature distribution of base classes using their abundant samples and estimate the feature distribution of each novel class according to their nearest base classes.
Finally, these methods sample more novel features from the estimated feature distribution and use them for training.

\noindent $\bullet$ \textbf{Sample Generation In the Feature Space $\rightarrow$ Generator-Based Methods.}
Halluc~\cite{zhang2021hallucination} aims to synthesize additional RoI features for novel classes.
It proposes a simple hallucinator to generate hallucinated RoI features, implemented as a simple two-layer MLP.
In the base-training stage, Halluc first trains a Faster R-CNN on the base dataset as regular object detection. Then, it freezes the parameters of the detector and pre-trains the hallucinator with a classification loss for the synthesized samples.
Next, in the few-shot fine-tuning stage, Halluc unfreezes the parameters of detection heads~(classification head \& regression head) and adopts an EM-like algorithm to train the hallucinator and detection heads alternately.
It is noted that this method shows impressive performance when the number of training samples is extremely small. However, its superiority over baseline methods such as TFA cannot be guaranteed as the number of training samples increases.

\noindent $\bigstar$ \textbf{Discussion of Data-Augmentation-Based Methods.}
Methods for sample generation in the input-pixel space are categorized into \textbf{multi-scale augmentation methods} and \textbf{novel instance mining methods}. Multi-scale augmentation methods are effective data-augmentation methods for FSOD, and they are easy to implement. However, conducting data-augmentation only on the aspect of scale does not tackle the core of data-scarcity problem of FSOD, and they are early FSOD methods with insufficient performance.
For the novel instance mining methods, it is true that on current FSOD benchmarks, many objects from novel classes indeed exist in the images without annotation. Capturing these objects effectively mitigates the data-scarcity problem in FSOD and significantly improves the performance. These methods have great potential to be integrated into other FSOD methods. However, this setting is not realistic. In real-life FSOD, it is not guaranteed that the images of the base dataset contain objects from novel classes.

Methods for sample generation in the feature space are categorized into two categories: \textbf{distribution inheritance methods} and \textbf{generator-based methods}.
The former type effectively generates more samples for novel classes using the data distribution from the data-abundant base classes. It introduces no extra parameters and can be considered as a ``free lunch'' from the base dataset. However, it is not applicable in the real-world scenario that there is a significant difference between the data distribution of the base classes and novel classes. The latter type is more suitable for the scenario that base classes and novel classes differ a lot, but it introduces an extra generator which may increase the burden for model training.

\subsubsection{\textbf{Pre-Train-Based Methods}}

Almost all transfer-learning methods adopt a backbone pre-trained on ImageNet before the base training stage.
Some methods (DETReg~\cite{bar2022CVPR}, imTED~\cite{zhang2022integral}) focus on improving this pre-training stage.

\begin{itemize}
    \item DETReg pre-trains a DETR model in an unsupervised manner. On the one hand, it uses Selective Search~\cite{uijlings2013selective} to excavate object proposals and uses them to train the object localization branch of the model. On the other hand, it uses another pre-trained self-supervised model to generate object encodings and enforces the DETR model to mimic these object encodings.
    \item imTED integrally migrates a pre-trained MAE model~\cite{he2022CVPR} to be a detection model. Concretely, imTED adds a region proposal network and a detection head into the MAE model following the design of Faster R-CNN. Besides, it proposes a multi-scale feature modulator to fuse multi-scale features extracted from a FPN~\cite{lin2017CVPR}.
\end{itemize}

\noindent $\bigstar$ \textbf{Discussion of Pre-Train-Based Methods.} These methods explore the current FSOD problem in a new perspective that pursues a stronger backbone before the few-shot training stage, while current FSOD methods most simply adopt a backbone pre-trained with a classification task on ImageNet. Besides, the performance of these methods is significantly superior to other methods. However, these methods require a stronger pre-trained backbone~(DETReg requires SwAV, and imTED requires MAE). Besides, these methods never clarify whether these stronger backbones cover the knowledge of novel classes in the FSOD setting, which will bring an unfair comparison with other FSOD methods.

\subsection{Fine-Tune-Free Methods}

Fine-tune-free methods focus on directly transferring the trained model from the base dataset to the novel dataset without fine-tuning.
Existing fine-tune-free methods~(AirDet~\cite{li2022ECCV}, FS-DETR~\cite{bulat2022fs}) adopt the scheme of meta-learning, and they also focus on the method of feature aggregation.
Specifically, AirDet conducts integral feature aggregation with element-wise multiplication and concatenation operations, and it proposes to learn the weights of different support samples instead of treating them as equals. Besides, AirDet aggregates RoI features with support features for the regression branch.
FS-DETR concatenates query features with support features into a common sequence and feeds it into the DETR model. FS-DETR proposes the learnable pseudo-class embeddings with the same shape as support features and adds them into support features to facilitate the model training.

\noindent $\bigstar$ \textbf{Discussion of Fine-Tune-Free Methods.}
Fine-tune-free method requires less computation cost and are more suitable for real life. However, the performance of these methods is currently not ideal compared to the fine-tune-based methods.

%% file: figures/fsod_meta_roi_aggregation.tex
\begin{figure}
\centering
\subfigure[FSRW~\cite{kang2019few} Meta R-CNN~\cite{yan2019meta}]{
    \begin{minipage}[t]{0.4\linewidth}
    \centering
    \includegraphics[width=4cm]{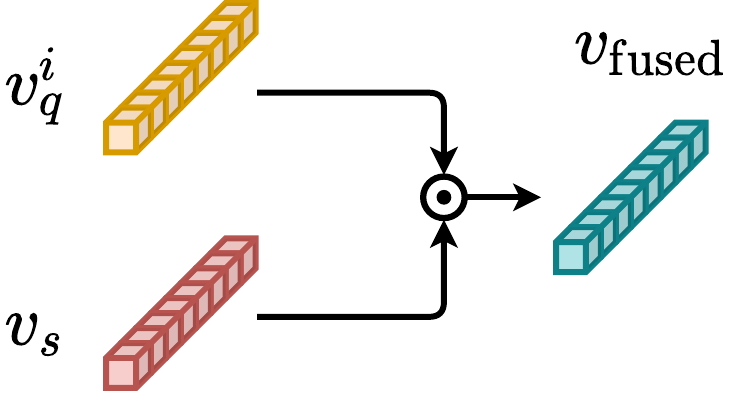}
    \label{fig:fsod_meta_roi_aggregation_fsrw}
    \end{minipage}%
}%
\subfigure[FsDet \cite{xiao2020few} AFD-Net \cite{liu2020afd}]{
    \begin{minipage}[t]{0.6\linewidth}
    \centering
    \includegraphics[width=7cm]{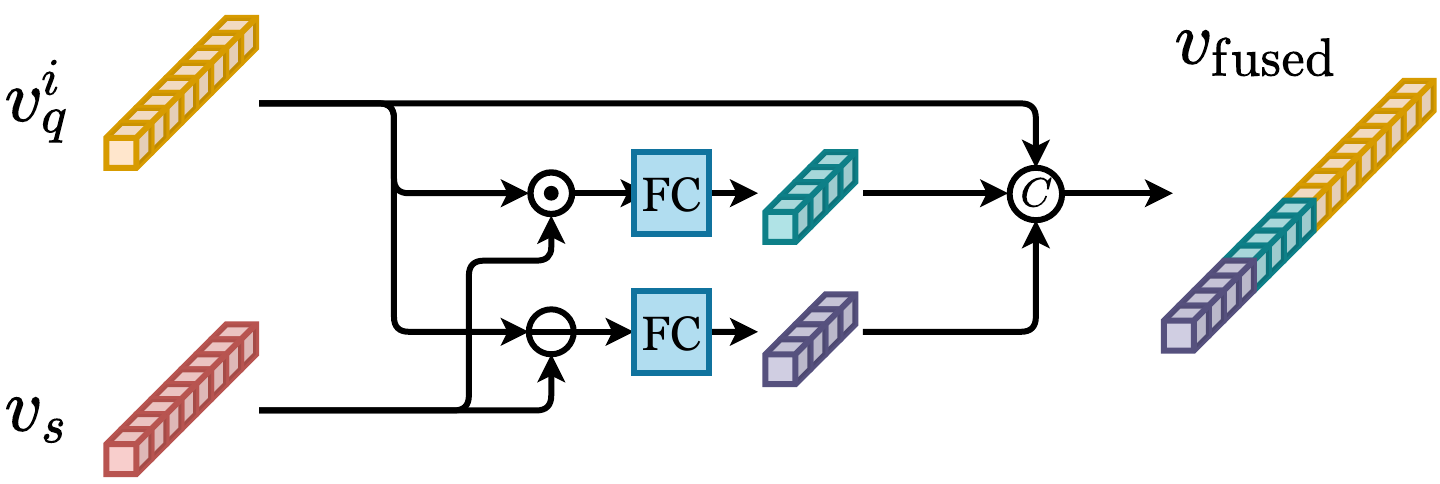}
    \label{fig:fsod_meta_roi_aggregation_fsdet}
    \end{minipage}%
}%
\centering
\caption{Overview of RoI feature vector concatenation method for standard FSOD. The symbols $\odot$, $\ominus$ and \textcircled{c} denotes element-wise multiplication, element-wise subtraction and concatenation operation, respectively. FSRW and Meta R-CNN aggregates the support feature vector and  the query feature vector with a simple element-wise multiplication. FsDet and AFD-Net concatenate the query feature vector with an element-wise multiplication, subtraction between the query feature vector and the support feature vector using two additional FC layers.}
\label{fig:fsod_meta_roi_aggregation}
\end{figure}

%% file: figures/fsod_transfer_learning.tex
\begin{figure}
\centering
\subfigure[Base Training Stage on Faster R-CNN]{
    \begin{minipage}[t]{0.8\linewidth}
    \centering
    \includegraphics[width=\linewidth]{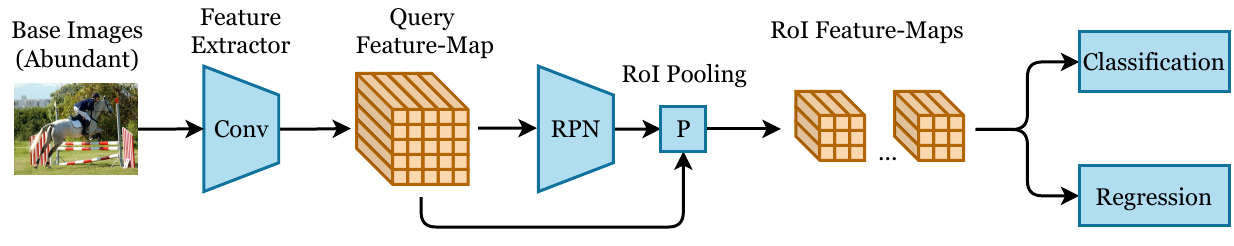}
    \end{minipage}%
}%

\subfigure[Few-Shot Fine-tuning Stage on Faster R-CNN framework]{
    \begin{minipage}[t]{0.8\linewidth}
    \centering
    \includegraphics[width=\linewidth]{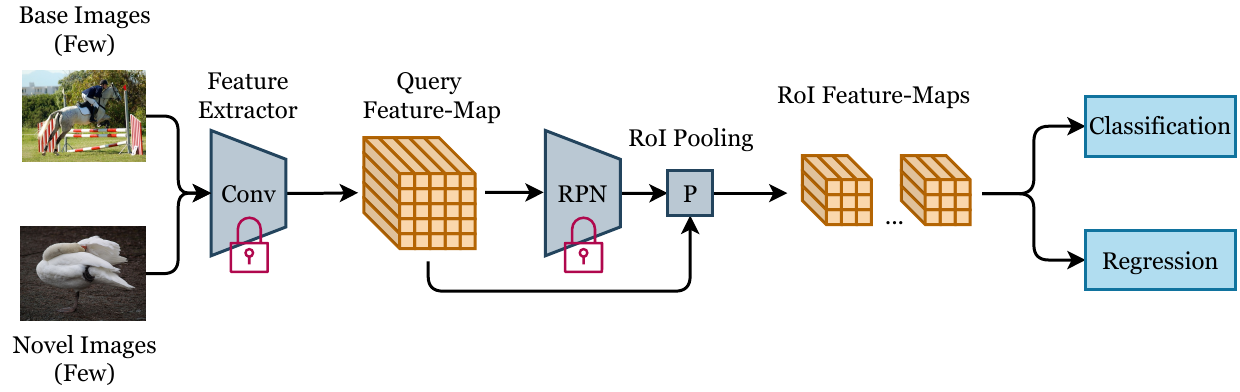}
    \end{minipage}%
}
\centering
\caption{Overview of the two-stage transfer-learning framework for standard FSOD \cite{wang2020few}. In the base training stage, the model is trained on the base dataset with abundant instances of base classes, while in the few-shot fine-tuning stage, the model is trained on a small dataset containing data for both base classes and novel classes. Current transfer-learning methods mostly adopt Faster R-CNN as the detection framework, as shown in the figure. The yellow components in these two figures denote intermediate tensors, the blue components denote modules in Faster R-CNN, and the lock symbol denotes that the parameters of the corresponding module are frozen.}
\label{fig:transfer_learning_fsod}
\end{figure}

%% file: n05_ZSOD.tex
\section{Zero-Shot Object Detection\label{sec:standard_zsd_methods}}

Zero-Shot Object Detection (ZSOD) is an extreme scenario of LSOD that novel classes do not contain any image sample. Concretely, the training dataset~(base dataset $D_B$) of ZSOD consists of abundant annotated instances of base classes $C_B$, and the test dataset~(novel dataset $D_N$) does not consist of annotated instances of novel classes $C_N$~($C_B$ and $C_N$ are not intersected). As a substitute, ZSOD utilizes semantic information to assist in detecting objects of novel classes.

According to whether utilizing unlabeled test images for model training, this survey categorizes ZSOD into two domains: ``transductive ZSOD'' and ``inductive ZSOD''. Inductive ZSOD is the mainstream of ZSOD, which does not require accessing the test images in advance.
Differently, transductive ZSOD is rarely explored, which utilizes unlabeled test images to assist model training. Furthermore, inductive ZSOD is categorized according to the type of semantic information: \textbf{semantic attributes} and \textbf{textual description}. The former type utilizes the semantic attributes~(word vector) as the auxiliary semantic information to represent each class. In contrast, the latter type utilizes the textual description~(e.g., a description sentence for an image or a class) as the auxiliary semantic information. This section gives a comprehensive introduction to semantic-attributes-based inductive ZSOD~(standard ZSOD). \textbf{Textual-description-based inductive ZSOD} and \textbf{transductive ZSOD} will be discussed in the later sections.

Current semantic-attributes-based inductive ZSOD methods adopt Faster R-CNN or YOLO-style model as the detection framework, as shown in \autoref{fig:zsd_inductive}. Ankan Bansal et al.~\cite{bansal2018zsd} propose one of the earliest methods for semantic-attributes-based inductive ZSOD based on Faster R-CNN.
This work first establishes a simple baseline built on Faster R-CNN, which uses a simple linear projection to project RoI features $v_r$ into semantic space and calculates the class probabilities of $v_r$ as the cosine similarities between the projected semantic embeddings $s_r$ and the semantic attributes of each class.
As one of the earliest methods for ZSOD, this work sets up a benchmark adopted by many future works.

\input{figures/ZSOD_framework}

\noindent $\bullet$ \textbf{ZS-YOLO}~\cite{zhu2018ZSD} is another early work for semantic-attributes-based inductive ZSOD based on YOLOv2.
It projects each cell in the feature map into semantic embeddings for class prediction.
Compared to the contemporaneous work~\cite{bansal2018zsd}, ZS-YOLO adopts a different detection framework, and it does not require external training data and semantic embeddings of background class. However, these two methods are evaluated using different dataset settings, making it difficult to directly compare their performance.

As the forerunners of two mainstream detection frameworks for semantic-attributes-based inductive ZSOD, the above two methods~\cite{bansal2018zsd, zhu2018ZSD} are followed by many future works.
The later methods mainly follow their framework with some extensions on different components of the framework. According to the modified components they focus on, this survey categorizes semantic-attributes-based inductive ZSOD methods into \textbf{semantic relation methods}, \textbf{data augmentation methods}, and \textbf{visual-semantic mapping methods}.

\subsection{Semantic Relation Methods}

Semantic relation methods utilize the semantic relation between classes to detect objects of novel classes, which are further categorized into \textbf{base-novel class relation} and \textbf{super-class relation}. Methods based on base-novel class relation utilize semantic similarities between base classes and novel classes to transfer knowledge from base classes to novel classes. Methods based on super-class relation assume that there is a hierarchical relationship among categories, i.e., some similar classes can be grouped into a super-class~(e.g., bed, sofa, and chair can be grouped into furniture), and they utilize this hierarchical relationship to assist prediction.

\subsubsection{\textbf{Base-Novel Class Relation}}

\input{figures/ZSOD_base_novel_relation}

Methods based on base-novel class relation can be categorized into two types: \textbf{linear-transform-based methods}~(TOPM-ZSOD~\cite{shao2019zero}, LSA-ZSOD~\cite{wang2020learning}, DPIF~\cite{li2021inference}), and \textbf{graph-based methods}~(SPGP~\cite{caixia2020semantics}, VSRG~\cite{nie2022WACV}, CRF-ZSOD~\cite{luo2020context_recognition}).
Linear-transform-based methods utilize the base-novel semantic relation to assist prediction through linear transforms of these semantic relations, and graph-based methods construct graphs with each node as a category, towards fully excavating the relation between base classes and novel classes through graph neural networks or conditional random fields.

\noindent $\bullet$ \textbf{Base-Novel Class Relation $\rightarrow$ Linear-Transform-Based Methods.} TOPM-ZSOD~\cite{shao2019zero}, LSA-ZSOD~\cite{wang2020learning}, and DPIF~\cite{li2021inference} are all linear-transform-based methods to utilize base-novel class relation for ZSOD.

\noindent $\bullet$ \textbf{Base-Novel Class Relation $\rightarrow$ Graph-Based Methods.} SPGP~\cite{caixia2020semantics}, VSRG~\cite{nie2022WACV}, and CRF-ZSOD~\cite{luo2020context_recognition} are graph-based methods to better excavate the relation between base and novel classes into the classification branch for higher performance.


\noindent $\bigstar$ \textbf{Discussion of Methods Based on Base-Novel Class Relation.} Linear-transform-based methods are simple approaches to directly utilize the base-novel semantic relation to assist prediction. However, linear transform does not fully excavate the base-novel relation for prediction, and it does not connect RoI features with the class semantic attributes together. Graph-based methods deeply excavate the relation between base classes and novel classes for prediction through graph neural networks or conditional random fields. Although they improve the performance through the graph structure modeling the relation between categories, they haven't provided a quantitative analysis of whether the trained graph matches human intuition.

\subsubsection{\textbf{Super-Class Relation}}

Methods based on super-class relation~(CG-ZSOD~\cite{li2020context}, JRLNC-ZSOD~\cite{rahman2020Joint}, ACS-ZSOD~\cite{mao2020zero}) define some coarse-grained classes~(super-classes) to cluster all classes into several groups, which separate the original classification problem into two sub-problems~(coarse-grained classification and fine-grained classification).


\noindent $\bigstar$ \textbf{Discussion of Methods Based on Super-Class Relation.} These methods provide ``free lunch'' for the performance improvement of ZSOD, but they are unsuitable for situation where there is no hierarchical relationship between categories.

\subsection{Visual-Semantic Mapping Methods}

Visual-semantic mapping methods aim to find a proper mapping function to align visual features with the class semantic attributes. Visual-semantic mapping methods can be categorized into \textbf{linear-projection-based methods}~(e.g., LSA-ZSOD~\cite{wang2020learning}, DPIF~\cite{li2021inference}, ZSDTR~\cite{zheng2021ICIP}), \textbf{weighted-combination-based methods}~(HRE-ZSOD~\cite{berkan2018hybrid}), \textbf{inverse-mapping methods}~(MS-ZSOD~\cite{gupta2020multispace}, CCFA-ZSOD~\cite{li2022zero}, SMFL-ZSOD~\cite{qianzhong2020rethinking}), \textbf{auxiliary-loss-based methods}~(ContrastZSOD~\cite{yan2022semantics}, VSA-ZSOD~\cite{rahman2020visual_semantic}), \textbf{external-resource-based methods}~(CLIP-ZSOD~\cite{xie2022ICDM}, BLC~\cite{ye2020background}).

\noindent $\bullet$ \textbf{Linear-Projection-Based Methods.}
The earliest ZSOD method~\cite{bansal2018zsd} adopts this simplest visual-semantic mapping method that projects visual features into semantic space through a linear projection, which is followed by many ZSOD methods~(e.g., LSA-ZSOD~\cite{wang2020learning}, DPIF~\cite{li2021inference}, ZSDTR~\cite{zheng2021ICIP}).
These methods are mostly based on CNN backbones, and only ZSDTR adopts DETR~\cite{NicolasDETR2020}~(a vision-transformer-based detector) which projects the proposal encodings into semantic space.

\noindent $\bullet$ \textbf{Weighted-Combination-Based Methods.}
HRE-ZSOD~\cite{berkan2018hybrid} calculates the semantic embeddings $s_r$ of the RoI feature $v_r$ as the weighted combination of different semantic attributes from all base classes $C_B$ according to their classification scores, as shown in \autoref{equa:zsod_v_s_mapping_weighted_combination}~($p_c$ denotes the probability that this RoI is predicted to be the base class $c$).
\begin{equation}
\label{equa:zsod_v_s_mapping_weighted_combination}
    s_r = \frac{1}{\sum\limits_{c \in C_B}p_c}\sum\limits_{c \in C_B} p_c s_c \text{,} 
\end{equation}

\noindent $\bullet$ \textbf{Inverse-Mapping Methods.}
Inverse-mapping methods~(MS-ZSOD~\cite{gupta2020multispace}, CCFA-ZSOD~\cite{li2022zero}, SMFL-ZSOD~\cite{qianzhong2020rethinking}) conversely project class semantic attributes into visual space to align the class semantic attributes with the visual features.

\noindent $\bullet$ \textbf{Auxiliary-Loss-Based methods.}
Auxiliary-loss-based methods~(ContrastZSOD~\cite{yan2022semantics}, VSA-ZSOD~\cite{rahman2020visual_semantic}) propose some auxiliary losses to facilitate the visual-semantic mapping.

\noindent $\bullet$ \textbf{External-Resource-Based Methods.}
These methods utilize external resources~(CLIP-ZSOD~\cite{xie2022ICDM}, BLC~\cite{ye2020background}) to better project visual features into semantic space.
Specifically, CLIP-ZSOD utilizes a strong pre-trained CLIP model~\cite{Radford2021ICML} for visual-semantic mapping, and BLC adopts external vocabulary for visual-semantic mapping.

\subsection{Data Augmentation Methods}

Data augmentation methods aim to generate multiple visual features for novel classes to mitigate the data-scarcity problem. The generated features are used to re-train the classifier of the detection model. Early data augmentation methods~(DELO~\cite{zhu2020don}) train a conditional generator with some auxiliary losses for data generalization, and later methods~(GTNet~\cite{zhao2020gtnet}, SYN-ZSOD~\cite{hayat2020synthesizing}, RSC-ZSOD~\cite{sarma2022resolving}, RRFS-ZSOD~\cite{huang2022CVPR}) all adopt GAN~(generative adversarial network).

\noindent $\bullet$ \textbf{DELO}~\cite{zhu2020don} adopts a conditional generator to synthesize visual features for novel classes. Specifically, the generator consists of an encoder to extract the latent features of the corresponding semantic embeddings, and a decoder to synthesize the visual features from the latent features. DELO adopts the conditional VAE loss to train this generator, including a KL divergence loss and a reconstruction loss. Besides, it proposes three additional losses to encourage the consistency between the reconstructed visual features and the original visual features.


\noindent $\bullet$ \textbf{GTNet}~\cite{zhao2020gtnet}, \textbf{SYN-ZSOD}~\cite{hayat2020synthesizing}, \textbf{RSC-ZSOD}~\cite{sarma2022resolving}, and \textbf{RRFS-ZSOD}~\cite{huang2022CVPR}. These methods all adopt GAN~(generative adversarial network) to generate visual features for novel classes. The GAN consists of a generator to synthesize visual features and a discriminator to determine whether the visual features are synthesized or not. These methods propose some elaborated extensions on this framework respectively.

\noindent $\bigstar$ \textbf{Discussion of Data Augmentation Methods.} Data augmentation methods directly tackle the data-scarcity problem in ZSOD in an intuitive way. Actually, data augmentation methods can be seen as the inverse of visual-semantic mapping methods~(i.e., mapping the class semantic attributes back into visual features). An important difference is that data augmentation methods incorporate intra-class variance into this mapping process, i.e., these methods generate different image features from different random noises for the same class.
However, these methods can only synthesize visual features instead of visual samples~(in the input pixel space), making it hard to interpret or visualize the synthesized samples. Besides, it is possible to substitute these methods by inverting the visual-semantic mapping functions.

\section{Extensional Zero-Shot Object Detection\label{sec:non_standard_zsd_methods}}

\subsection{Open-Vocabulary Object Detection}

Conventional ZSOD only learns to align visual features with semantic information for detection from a small set of base classes~($C_B$) and generalizes to the novel classes~($C_N$), while Open-Vocabulary Object Detection~(OVD) first accesses a much larger dataset~(consisting of massive image-text pairs from multiple classes $C_O$) to train a stronger visual-semantic mapping function for multiple classes~(intersecting with the base and novel classes for the later ZSOD task).
We provide a detailed analysis of OVD in section {\bf S3} of the supplementary online-only material.


\subsection{Textual-Description-Based Inductive ZSOD}

Previous ZSOD methods use semantic attributes as semantic information to represent each class. Instead, textual-description-based methods use textual description as semantic information. Currently, only a few methods uncover textual-description-based inductive ZSOD, and they use different types of textual-description: \textbf{class textual description}~(description text for each class) and \textbf{image textual description}~(description text for each image).

\noindent $\bullet$ \textbf{Methods Based on Class Textual Description.}
ZSOD-TD~\cite{li2019textual} adopts textual description to represent each class instead of semantic attributes~(e.g., ``stripe, equid" is used to describe zebra).
ZSOD-TD projects the RoI features into semantic embeddings and makes predictions by comparing them with the features extracted from textual description.

\noindent $\bullet$ \textbf{Methods Based on Image Textual Description.}
In addition to the class textual description, ZSOD-CNN~\cite{zhang2020textual_cnn} adopts textual description to represent each image~(e.g., ``A bathroom with a sink and three towels."), which also adopts Faster R-CNN as the detection framework.
It uses a text CNN to extract text features, and concatenates the RoI features with the text features for further predictions.
Besides, this method utilizes the OHEM technique to select hard samples for model training. During testing, it predicts the classification scores of novel classes according to those of base classes according to the semantic similarities between base and novel classes.

\subsection{Transductive ZSOD}

\noindent $\bullet$ \textbf{Transductive ZSOD}~\cite{rahman2019transductive}. Transductive ZSOD is an extended setting of inductive ZSOD, which incorporates unlabeled test images into model training.
Rahman et al.~\cite{rahman2019transductive} propose the first work to uncover transductive ZSOD, which conducts transductive learning on a pre-trained ZSOD model.
For transductive learning, it applies a pseudo-labeling paradigm on the unlabeled data, including a fixed pseudo-labeling step to generate fixed pseudo-labels for base classes using the pre-trained model, and a dynamic pseudo-labeling step to generate pseudo-labels for both base classes and novel classes iteratively.
This work is the first to explore transductive learning on ZSOD, which shows promising potential for significant performance improvement, as other transductive methods in few-shot image classification.

%% file: figures/ZSOD_framework.tex
\begin{figure}
\centering
\subfigure[ZSOD model based on Faster R-CNN]{
    \begin{minipage}[t]{0.9\linewidth}
    \centering
    \includegraphics[width=\linewidth]{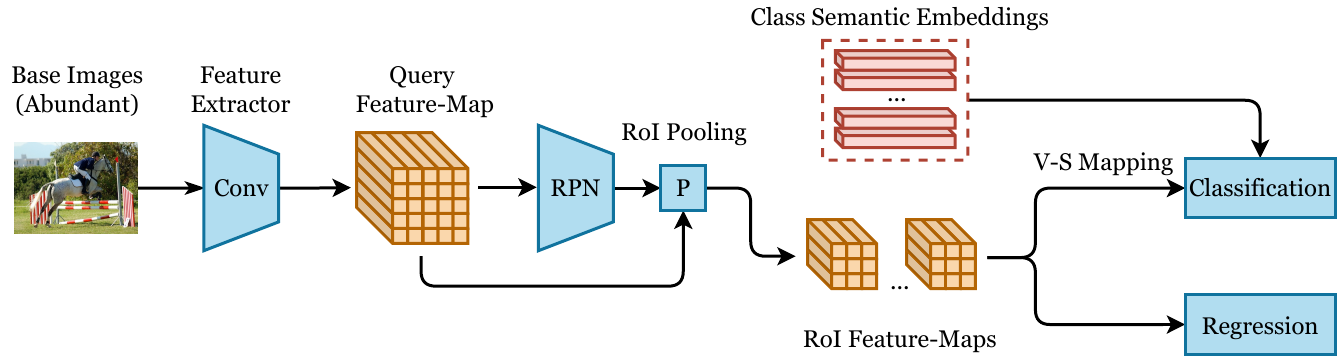}
    \end{minipage}%
}%

\subfigure[ZSOD model based on YOLO-style model]{
    \begin{minipage}[t]{0.8\linewidth}
    \centering
    \includegraphics[width=\linewidth]{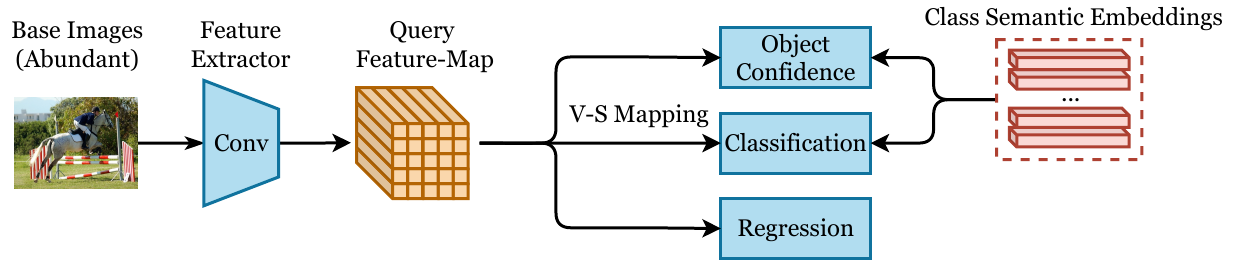}
    \end{minipage}%
}
\centering
\caption{Overview of two detection frameworks for ZSOD methods: Faster R-CNN and YOLO-style model. Most of the current methods apply a visual-semantic mapping operation to project visual features into semantic space and compare these projected semantic embeddings with class semantic embeddings for classification.}
\label{fig:zsd_inductive}
\end{figure}

%% file: figures/ZSOD_base_novel_relation.tex
\begin{figure}
\centering
\includegraphics[width=\textwidth]{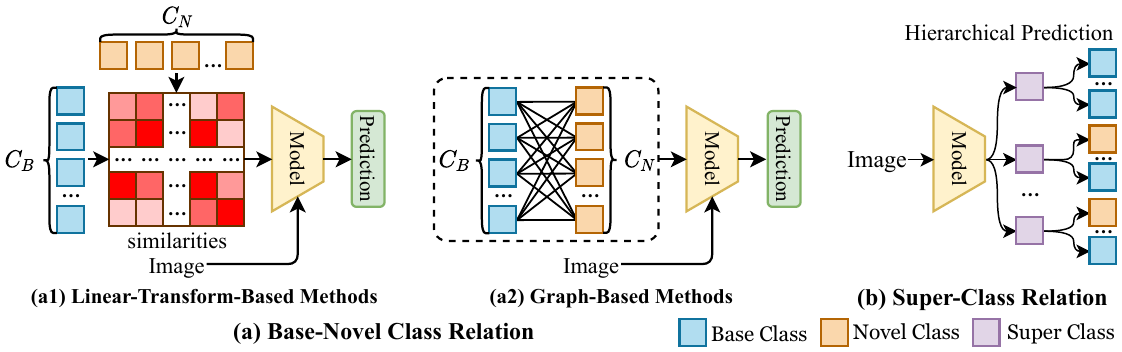}\\
\caption{Illustration of semantic relation methods: \textbf{base-novel class relation} and \textbf{super-class relation}.}
\label{fig:zsod_base_novel_super}
\end{figure}

%% file: n06_benchmarks.tex
\section{Popular Benchmarks For Low-Shot Object Detection\label{sec:dataset_eval}}

\subsection{Dataset Overview}

In three settings (i.e., OSOL, FSOD, and ZSOD) of LSOD, the classes of the dataset are all split into two types: base classes with large labeled samples and novel classes with few or no labeled samples. The mainstream benchmarks for Low-Shot Object Detection are modified from widely-used object detection datasets like the PASCAL VOC dataset, MS COCO dataset. This survey summarizes the basic information of mainstream benchmarks for LSOD in \autoref{tab:sum_low_shot_benchmark} but omits some rarely-used benchmarks since they are not representative. In this table, the number of base classes, the number of novel classes, and the number of labeled samples per category for each benchmark are recorded. Moreover, \textbf{split number} denotes the number of category split schemes for each benchmark.

\subsection{Evaluation Criteria}

\noindent \textbf{OSOL.} OSOL has a guarantee that the model knows precisely the object classes contained in each test image. For each test image in the test stage, OSOL randomly samples one support image for each category existing in this image to locate the objects of this category and average their accuracy scores as the final results.

\noindent \textbf{FSOD.} Different from OSOL, FSOD methods randomly sample a small set of support samples for the whole test set instead of only one image. For the K-shot setting, some methods like LSTD~\cite{chen2018lstd} sample K support images for each novel category. This sampling strategy is not ideal since the number of objects in the images may differ. Current methods mostly sample K bounding boxes for each novel category instead, and this survey records the performance of FSOD methods under this setting. Early FSOD methods mostly adopt the support samples released by FSRW~\cite{kang2019few} for fair performance comparison, which are sampled only once. TFA~\cite{wang2020few} samples support samples multiple times to obtain the average performance of the model. Currently, newly proposed FSOD methods mostly adopt this multiple sampling strategy to obtain more accurate performance.

\noindent \textbf{ZSOD.} ZSOD methods adopt two evaluation criteria for model performance comparison. The first criterion evaluates the model on a subset of test data that contains only objects of novel classes~(ZSOD). The second setting, generalized ZSOD~(GZSOD), evaluates the model on the complete test data, requiring the model to detect objects of both base classes and novel classes. Generalized ZSOD separately computes the mean average precision and recall of base classes and novel classes and uses a harmonic average to generate the average performance.

It is noted that the class semantic attributes for ZSOD are mainly borrowed from pre-trained word vectors or manually designed attributes: GloVe~(300-dim)~\cite{Jeffrey2014Glove}, BERT~(768-dim)~\cite{Jacob2019BERT}, word2vec~(300-dim)~\cite{Tom2013Word2vec}, fastText~\cite{Piotr2017fastText} and aPaY~(64-dim)~\cite{Ali2009aPaY}. Among them, aPaY contains manually designed attributes, and others contain pre-trained word vectors.

\begin{table}
\footnotesize
\centering
\caption{Summation Of Mainstream Benchmarks for Low-Shot Object Detection}
\label{tab:sum_low_shot_benchmark}
\setlength{\tabcolsep}{1.8mm}{%
\begin{tabular}{cccccc} 
    \toprule
   \textbf{LSOD Type} & \textbf{Dataset} & \textbf{Base Classes} & \textbf{Novel Classes} & \textbf{Shots Per Category} & \textbf{Split Number}
  \\
  \midrule
  \multirow{2} * {OSOL} & PASCAL VOC Dataset & 16 & 4 & 1 & 1 \\
  ~ & MS COCO Dataset & 60 & 20 & 1 & 4 \\
  \midrule
  \multirow{2} * {FSOD} & PASCAL VOC Dataset & 15 & 5 & 1, 2, 3, 5, 10 & 3 \\
  ~ & MS COCO Dataset & 60 & 20 & 10, 30 & 1 \\
  \midrule
  \multirow{3} * {ZSOD} & PASCAL VOC Dataset & 16 & 4 & 0 & 1 \\
  ~ & MS COCO Dataset & 48 & 17 & 0 & 1 \\
  ~ & MS COCO Dataset & 65 & 15 & 0 & 1 \\
  \bottomrule
  \end{tabular}}
\end{table}

\subsection{Evaluation Metrics}

\noindent $\bullet$ Preliminaries for the calculation of evaluation metrics:

\textbf{Intersection over Union~(IoU).} Intersection over Union~(IoU) is a value that measures the degree of overlap between two bounding boxes. Specifically, let $\mathrm{bbox}_{1} \cap \mathrm{bbox}_{2}$ and $\mathrm{bbox}_{1} \cup \mathrm{bbox}_{2}$ respectively denote the area of overlap and union of two bounding boxes $\mathrm{bbox}_{1}$ and $\mathrm{bbox}_{2}$, the IoU between them $\mathrm{IoU}(\mathrm{bbox}_{1}, \mathrm{bbox}_{2})$ is calculated as $\mathrm{IoU}(\mathrm{bbox}_{1}, \mathrm{bbox}_{2}) = \frac{\mathrm{bbox}_{1} \ \cap \ \mathrm{bbox}_{2}}{\mathrm{bbox}_{1} \ \cup \ \mathrm{bbox}_{2}}$. Two bounding boxes are considered to be matched if their IoU is larger than a pre-determined threshold $\mu$.

\noindent $\bullet$ The evaluation metrics for LSOD:

\textbf{Precision.} Precision is the fraction of correctly retrieved bounding boxes out of total retrieved bounding boxes.


\textbf{Recall@K.} In converse to Precision, Recall is the fraction of correctly retrieved bounding boxes out of total ground-truth bounding boxes~(\textbf{K} denotes the number of total retrieved bounding boxes).


\textbf{mAP50.} AP50~(average precision with $\mu = 0.5$) is the precision averaged over different levels of recall. Let $\mathrm{Prec}\ ({\mathrm{recall\_value}})$ denote the precision when ``$\mathrm{recall\_value}$'' is achieved, and AP50 is calculated averaged over some specific values $\mathcal{R}$ of recall~($\mathcal{R} = \{ 0, 0.1, 0.2, ... , 1.0 \}$ is usually selected), as shown in \autoref{equa:ap50_cal}.
AP50 is calculated for each category and their results are averaged as the final mAP50~(mean average precision with $\mu = 0.5$). Note that mAP50 is commonly adopted on the PASCAL VOC benchmark.

\begin{equation}
    \label{equa:ap50_cal}
    \mathrm{AP50} = \frac{1}{| \mathcal{R} |} \sum\limits_{\mathrm{recall\_value} \in \mathcal{R}}\mathrm{Prec} \ (\mathrm{recall\_value}).
\end{equation}

\textbf{mAP.} mAP is the extension of mAP50 that is averaged over ten IoU thresholds: $\{0.5, 0.55, 0.60, ... \ , \\ 0.95 \}$, which is commonly adopted on the MS COCO benchmark.

\section{Performance\label{sec:performance}}

This section demonstrates and analyzes the performance of different Low-Shot Object Detection methods on the most widely-used benchmarks.

\subsection{One-Shot Object Localization}

\autoref{tab:osod_performance} lists the performance of current OSOL methods on the PASCAL VOC benchmark and the MS COCO benchmark~(the results on the MS COCO benchmark are averaged over $4$ splits). SiamFC and SiamRPN are two methods initially proposed for video object tracking, which are the baselines for OSOL, and their performance is reasonably poor than authentic OSOL methods.
SiamMask, OSCD, OSOLwT, and FOC OSOL use simple concatenation-based methods for feature aggregation with different modifications. These methods significantly outperform SiamFC \& SiamRPN, but they have performance inferior to the attention-based methods, and FOC OSOL achieves the best performance among these methods on the PASCAL VOC benchmark.
Differently, recently proposed methods~(CoAE, ADA OSOL, AUG OSOL, AIT, CAT, BHRL, SaFT, ABA OSOL) most adopt the attention mechanism for feature aggregation, and CAT is the best method among them. Moreover, CAT~(a transformer-based method) achieves $4.5$ points better than FOC OSOL on the PASCAL VOC benchmark, which indicates that attention-based methods are more promising for future One-Shot Object Localization.

\input{tables/osol_benchmark}

\subsection{Few-Shot Object Detection}

This subsection demonstrates the performance of standard Few-Shot Object Detection methods on two most commonly used benchmarks: PASCAL VOC benchmark and MS COCO benchmark. For a fair comparison, this survey only lists the performance of FSOD methods \textbf{with released codes}.

\autoref{tab:pascalvoc_performance}, \autoref{tab:pascalvoc_avg_performance} and \autoref{tab:coco_performance} present the performance on novel classes of PASCAL VOC benchmark and MS COCO benchmark, respectively. Some conclusions can be summarized from these two tables:
(1) The best-performing transfer-learning method have superior performance to the best performing meta-learning methods on the most commonly used backbone~(ResNet-101). Specifically, FSOD-DIS~(the best-performing transfer-learning method on ResNet-101) exceeds VFA~(the best-performing meta-learning method on ResNet-101) on the MS COCO benchmark.
(2) For meta-learning methods, mixed feature aggregation methods outperform RoI feature aggregation methods on two benchmarks overall. The reasons for this phenomenon is that mixed feature aggregation methods incorporate category-specific information into the shallow components~(RPN, mainly) of the detection model, which directly guides the prediction of these components using the support information.
(3) For transfer-learning methods, data augmentation methods~(e.g., Halluc, PSEUDO, FSOD-DIS) show strong performance in an extremely few-shot condition~(${\rm shot} = 1, 2, 3$), demonstrating that data augmentation methods effectively tackle the data-scarcity problem in the extremely few-shot condition.
(4) Methods on advanced backbones~(FCT on PVTv2-B2-Li, Meta-DETR \& DETReg on Def. DETR, PSEUDO on Swin-S, imTED on ViT-B) show significantly higher performance than methods on the regular backbone~(ResNet-50 \& ResNet-101), which point out a promising direction for the development of FSOD.
(5) The performance ranking of a method can differ across these two benchmarks.

\input{tables/fsod_pascal_benchmark}
\input{tables/fsod_pascal_avg_benchmark}

\subsection{Zero-Shot Object Detection}

\autoref{tab:gzsd_coco_benchmark_4817} and \autoref{tab:gzsd_coco_benchmark_6515} demonstrate the performance of standard ZSOD methods under two evaluation protocols~(ZSOD, GZSOD) on the most commonly used benchmark: MS COCO benchmark. Some trends can be found in this table.
(1) Early ZSOD methods are not consistent in the choice of semantic attributes, and only a few of them are evaluated under the GZSOD protocol. Nevertheless, the newly proposed ZSOD methods mostly adopt word2vec as their semantic attributes and use both ZSOD protocol and GZSOD protocol to evaluate the model, which is more convenient for performance comparison.
(2) The model performance of $48/17$ base-novel split is generally inferior to that of $65/15$ base-novel split, which is attributed to the fewer classes and samples in the base dataset.
(3) Current data augmentation methods for ZSOD cannot achieve satisfying performance compared to the newly proposed ZSOD methods. However, data augmentation methods can outperform other methods when the shot number is small in FSOD, which is still promising in ZSOD.
(4) The newly proposed ZSOD methods, such as CLIP-ZSOD, incorporate pre-trained cross-modal models like CLIP in their training process and achieve remarkable performance compared to state-of-the-art methods. This demonstrates the potential to transfer external foundation models in future ZSOD research, leading to even higher performance.

%% file: tables/osol_benchmark.tex
\begin{table}
\footnotesize
\renewcommand\arraystretch{1.1}
\centering
\caption{Performance~(mAP50) of OSOL methods on novel classes. On each benchmark, the red font denotes the best performance, and the gray font denotes the second-best performance. Note that PASCAL VOC has only one class split, while the results on MS COCO are averaged over four different class splits. R-50 \& R-101 denotes ResNet-50 \& ResNet-101.}
\label{tab:osod_performance}
\setlength{\tabcolsep}{1.mm}{
\begin{tabular}{ c
*4{c}} 
    \toprule
  \textbf{Type} & \textbf{Method} & \textbf{Detector~(Backbone)} & \textbf{PASCAL VOC} & \textbf{MS COCO} \\
  \midrule
\multirow{2}*{Object Tracking Methods} & SiamFC~(2018)~\cite{Cen2018SiamFC} & SNet \& ENet~(VGG-16) & 13.3 & N/A \\
~ & SiamRPN~(2018)~\cite{Li2018SiamRPN} & Faster R-CNN~(AlexNet) & 14.2 & N/A \\

\hline

\multirow{4}*{Concatenation-Based Methods} & SiamMask~(2019)~\cite{Ivan2018SiamMask} & Faster R-CNN~(R-50) & N/A & 16.8 \\

~ & OSCD~(2020)~\cite{FuZ0S21} & Faster R-CNN~(AlexNet) & 52.1 & N/A \\

~ & OSOLwT~(2020)~\cite{Li2021OSODwF} & Faster R-CNN~(R-50) & 69.1 & N/A \\

~ & FOC OSOL~(2021)~\cite{Yang2021foc} & Faster R-CNN~(R-50) & 71.0 & N/A \\

\hline

\multirow{8}*{Attention-Based Methods} & CoAE~(2019)~\cite{HsiehLCL19} & Faster R-CNN~(R-50) & 68.2 & 22.0 \\

~ & ADA OSOL~(2022)~\cite{zhang2022neuro} & Faster R-CNN~(R-50) & 72.3 & 23.6 \\

~ & AUG OSOL~(2022)~\cite{du2022neuro} & Faster R-CNN~(R-50) & 73.2 & 23.9 \\

~ & AIT~(2021)~\cite{Chen2021AIT} & Faster R-CNN~(R-50) & 73.1 & 24.3 \\

~ & BHRL~(2022)~\cite{yang2022CVPR} & Faster R-CNN~(R-50) & 73.8 & \textbf{\color{red} 25.6} \\

~ & SaFT~(2022)~\cite{zhao2022CVPR} & FCOS~\cite{tian2019FSOS}~(R-101) & 74.5 & {\color{gray} 24.9} \\

~ & CAT~(2021)~\cite{weidongCAT2021} & Faster R-CNN~(R-50) & \textbf{\color{red} 75.5} & 24.4 \\

~ & ABA OSOL~(2023)~\cite{hsieh2023WACV} & Faster R-CNN~(R-50) & {\color{gray} 74.6} & 23.6 \\

\bottomrule
\end{tabular}}
\end{table}

%% file: tables/fsod_pascal_benchmark.tex
\begin{table}
\footnotesize
\renewcommand\arraystretch{1.1}
\centering
\caption{Performance~(mAP50) of FSOD methods on the PASCAL VOC benchmark~(only the methods with released codes are listed). These FSOD methods are evaluated on the three splits of PASCAL VOC dataset under the $1, 2, 3, 5, 10$-shot condition. For each shot, the red font denotes the best performance, and the gray font denotes the second-best performance. ${}^{\star}$ denotes that the results are averaged over multiple runs, and R-101 denotes ResNet-101.}
\label{tab:pascalvoc_performance}
\setlength{\tabcolsep}{0.25mm}{
\begin{tabular}{c|c|c|
*5{c}|*5{c}|*5{c}} 
    \toprule
  \multirow{2}*{} & \multirow{2}*{\textbf{Method}} & \multirow{2}*{\textbf{Detector~(Backbone)}} & \multicolumn{5}{c|}{\textbf{Novel Set 1}} &  \multicolumn{5}{c|}{\textbf{Novel Set 2}} & \multicolumn{5}{c}{\textbf{Novel Set 3}} \\
  ~ & ~ & ~ & \textbf{1} & \textbf{2} & \textbf{3} & \textbf{5} & \textbf{10} & \textbf{1} & \textbf{2} & \textbf{3} & \textbf{5} & \textbf{10} & \textbf{1} & \textbf{2} & \textbf{3} & \textbf{5} & \textbf{10} \\
  \midrule

\multirow{10}*{\rotatebox{270}{\textbf{Meta-Learning}}} & FSRW~(2018) & YOLOv2 & 14.8 & 15.5 & 26.7 & 33.9 & 47.2 & 15.7 & 15.3 & 22.7 & 30.1 & 40.5 & 21.3 & 25.6 & 28.4 & 42.8 & 45.9 \\

~ & Meta-RCNN~(2019) & Faster R-CNN~(R-101) & 19.9 & 25.5 & 35.0 & 45.7 & 51.5 & 10.4 & 19.4 & 29.6 & 34.8 & 45.4 & 14.3 & 18.2 & 27.5 & 41.2 & 48.1 \\

~ & FsDet~(2020)${}^{\star}$ & Faster R-CNN~(R-101) & 24.2 & 35.3 & 42.2 & 49.1 & 57.4 & 21.6 & 24.6 & 31.9 & 37.0 & 45.7 & 21.2 & 30.0 & 37.2 & 43.8 & 49.6 \\

~ & DRL~(2021)${}^{\star}$ & Faster R-CNN~(R-101) & 30.3 & 40.8 & 49.1 & 48.0 & 58.6 & 22.4 & 36.1 & 36.9 & 35.4 & 51.8 & 24.8 & 29.3 & 37.9 & 43.6 & 50.4 \\

~ & DCNet~(2021)${}^{\star}$ & Faster R-CNN~(R-101) & 33.9 & 37.4 & 43.7 & 51.1 & 59.6 & 23.2 & 24.8 & 30.6 & 36.7 & 46.6 & 32.3 & 34.9 & 39.7 & 42.6 & 50.7 \\

~ & CME~(2021) & YOLOv2 & 17.8 & 26.1 & 31.5 & 44.8 & 47.5 & 12.7 & 17.4 & 27.1 & 33.7 & 40.0 & 15.7 & 27.4 & 30.7 & 44.9 & 48.8 \\

~ & Meta-DETR~(2022)${}^{\star}$ & Def. DETR~(R-101) & 35.1 & 49.0 & 53.2 & 57.4 & 62.0 & 27.9 & 32.3 & 38.4 & 43.2 & 51.8 & 34.9 & 41.8 & 47.1 & 54.1 & 58.2 \\

~ & QA-FewDet~(2021) & Faster R-CNN~(R-101) & 42.4 & 51.9 & 55.7 & 62.6 & 63.4 & 25.9 & 37.8 & 46.6 & 48.9 & 51.1 & 35.2 & 42.9 & 47.8 & 54.8 & 53.5 \\


~ & FCT~(2022)${}^{\star}$ & {\scriptsize Faster R-CNN~(PVTv2-B2-Li)} & 38.5 & 49.6 & 53.5 & 59.8 & 64.3 & 25.9 & 34.2 & 40.1 & 44.9 & 47.4 & 34.7 & 43.9 & 49.3 & 53.1 & 56.3 \\


~ & VFA~(2023) & Faster R-CNN~(R-101) & {\color{gray} 57.7} & {\color{gray} 64.6} & {\color{gray} 64.7} & {\color{gray} 67.2} & 67.4 & 41.4 & {\color{gray} 46.2} & {\color{gray} 51.1} & 51.8 & 51.6 & 48.9 & 54.8 & 56.6 & 59.0 & 58.9 \\
\hline


\multirow{11}*{\rotatebox{270}{\textbf{Transfer-Learning}}} & TFA w/cos~(2020) & Faster R-CNN~(R-101) & 39.8 & 36.1 & 44.7 & 55.7 & 56.0 & 23.5 & 26.9 & 34.1 & 35.1 & 39.1 & 30.8 & 34.8 & 42.8 & 49.5 & 49.8 \\


~ & Halluc~(2021) & Faster R-CNN~(R-101) & 47.0 & 44.9 & 46.5 & 54.7 & 54.7 & 26.3 & 31.8 & 37.4 & 37.4 & 41.2 & 40.4 & 42.1 & 43.3 & 51.4 & 49.6 \\

~ & MPSR~(2020) & Faster R-CNN~(R-101) & 41.7 & N/A & 51.4 & 55.2 & 61.8 & 24.4 & N/A & 39.2 & 39.9 & 47.8 & 35.6 & N/A & 42.3 & 48.0 & 49.7 \\

~ & ${\rm FSOD}^{\rm up}(2021)$ & Faster R-CNN~(R-101) & 43.8 & 47.8 & 50.3 & 55.4 & 61.7 & 31.2 & 30.5 & 41.2 & 42.2 & 48.3 & 35.5 & 39.7 & 43.9 & 50.6 & 53.5 \\

~ & FSCE~(2021)${}^{\star}$ & Faster R-CNN~(R-101) & 32.9 & 44.0 & 46.8 & 52.9 & 59.7 & 23.7 & 30.6 & 38.4 & 46.0 & 48.5 & 22.6 & 33.4 & 39.5 & 47.3 & 54.0 \\

~ & DeFRCN~(2021)${}^{\star}$ & Faster R-CNN~(R-101) & 40.2 & 53.6 & 58.2 & 63.6 & 66.5 & 29.5 & 39.7 & 43.4 & 48.1 & {\color{gray} 52.8} & 35.0 & 38.3 & 52.9 & 57.7 & 60.8 \\

~ & FSOD-KI~(2022) & Faster R-CNN~(R-101) & 57.0 & 62.3 & 63.3 & 66.2 & {\color{gray} 67.6} & \textbf{\color{red} 42.8} & 44.9 & 50.5 & {\color{gray} 52.3} & 52.2 & {\color{gray} 50.8} & {\color{gray} 56.9} & {\color{gray} 58.5} & {\color{gray} 62.1} & {\color{gray} 63.1} \\

~ & FSOD-KD~(2022) & Faster R-CNN~(R-101) & 46.7 & 53.1 & 53.8 & 61.0 & 62.1 & 30.1 & 34.2 & 41.6 & 41.9 & 44.8 & 41.0 & 46.0 & 47.2 & 55.4 & 55.6 \\

~ & FADI~(2022) & Faster R-CNN~(R-101) & 50.3 & 54.8 & 54.2 & 59.3 & 63.2 & 30.6 & 35.0 & 40.3 & 42.8 & 48.0 & 45.7 & 49.7 & 49.1 & 55.0 & 59.6 \\

~ & PSEUDO~(2022) & Faster R-CNN~(R-101) & 54.5 & 53.2 & 58.8 & 63.2 & 65.7 & 32.8 & 29.2 & 50.7 & 49.8 & 50.6 & 48.4 & 52.7 & 55.0 & 59.6 & 59.6 \\

~ & FSOD-DIS~(2022) & Faster R-CNN~(R-101) & \textbf{\color{red} 63.4} & \textbf{\color{red} 66.3} & \textbf{\color{red} 67.7} & \textbf{\color{red} 69.4} & \textbf{\color{red} 68.1} & {\color{gray} 42.1} & \textbf{\color{red} 46.5} & \textbf{\color{red} 53.4} & \textbf{\color{red} 55.3} & \textbf{\color{red} 53.8} & \textbf{\color{red} 56.1} & \textbf{\color{red} 58.3} & \textbf{\color{red} 59.0} & \textbf{\color{red} 62.2} & \textbf{\color{red} 63.7} \\
\bottomrule
\end{tabular}}
\end{table}

%% file: tables/fsod_pascal_avg_benchmark.tex
\begin{table}
\footnotesize
\renewcommand\arraystretch{1.1}
\centering
\caption{Performance~(mAP50) of FSOD methods on the PASCAL VOC benchmark~(only the methods with released codes are listed). These FSOD methods are evaluated on the PASCAL VOC dataset under the $1, 2, 3, 5, 10$-shot condition. The results are averaged over three splits of base \& novel classes. For each shot, the red font denotes the best performance, and the gray font denotes the second-best performance. ${}^{\star}$ denotes that the results are averaged over multiple runs, and R-101 denotes ResNet-101.}
\label{tab:pascalvoc_avg_performance}
\setlength{\tabcolsep}{0.8mm}{
\begin{tabular}{c|c|c|
*5{c}} 
    \toprule
  \multirow{2}*{} & \multirow{2}*{\textbf{Method}} & \multirow{2}*{\textbf{Detector~(Backbone)}} & \multicolumn{5}{c}{\textbf{3 Novel Sets~(Averaged)}} \\
  ~ & ~ & ~ & \textbf{1} & \textbf{2} & \textbf{3} & \textbf{5} & \textbf{10} \\
  \midrule

\multirow{10}*{\rotatebox{270}{\textbf{Meta-Learning}}}  &  FSRW~(2018)  &  YOLOv2  & 17.3 & 18.8 & 25.9 & 35.6 & 44.5 \\
~  &  Meta-RCNN~(2019)  &  Faster R-CNN~(R-101)  & 14.9 & 21.0 & 30.7 & 40.6 & 48.3 \\
~  &  FsDet~(2020)${}^{\star}$  &  Faster R-CNN~(R-101)  & 22.3 & 30.0 & 37.1 & 43.3 & 50.9 \\
~  &  DRL~(2021)${}^{\star}$  &  Faster R-CNN~(R-101)  & 25.8 & 35.4 & 41.3 & 42.3 & 53.6 \\
~  &  DCNet~(2021)${}^{\star}$  &  Faster R-CNN~(R-101)  & 29.8 & 32.4 & 38.0 & 43.5 & 52.3 \\
~  &  CME~(2021)  &  YOLOv2  & 15.4 & 23.6 & 29.8 & 41.1 & 45.4 \\
~  &  Meta-DETR~(2022)${}^{\star}$  &  Def. DETR~(R-101)  & 32.6 & 41.0 & 46.2 & 51.6 & 57.3 \\
~  &  QA-FewDet~(2021)  &  Faster R-CNN~(R-101)  & 34.5 & 44.2 & 50.0 & 55.4 & 56.0 \\
~  &  FCT~(2022)${}^{\star}$  &  {\scriptsize Faster R-CNN~(PVTv2-B2-Li)}  & 33.0 & 42.6 & 47.6 & 52.6 & 56.0 \\
~  &  VFA~(2023)  &  Faster R-CNN~(R-101)  & 49.3 & {\color{gray} 55.2} & {\color{gray} 57.5} & 59.3 & 59.3 \\
\hline
\multirow{11}*{\rotatebox{270}{\textbf{Transfer-Learning}}}  &  TFA w/cos~(2020)  &  Faster R-CNN~(R-101)  & 31.4 & 32.6 & 40.5 & 46.8 & 48.3 \\
~  &  Halluc~(2021)  &  Faster R-CNN~(R-101)  & 37.9 & 39.6 & 42.4 & 47.8 & 48.5 \\
~  &  MPSR~(2020)  &  Faster R-CNN~(R-101)  & 33.9 & N/A & 44.3 & 47.7 & 53.1 \\
~  &  ${\rm FSOD}^{\rm up}(2021)$  &  Faster R-CNN~(R-101)  & 36.8 & 39.3 & 45.1 & 49.4 & 54.5 \\
~  &  FSCE~(2021)${}^{\star}$  &  Faster R-CNN~(R-101)  & 26.4 & 36.0 & 41.6 & 48.7 & 54.1 \\
~  &  DeFRCN~(2021)${}^{\star}$  &  Faster R-CNN~(R-101)  & 34.9 & 43.9 & 51.5 & 56.5 & 60.0 \\
~  &  FSOD-KI~(2022)  &  Faster R-CNN~(R-101)  & {\color{gray} 50.2} & 54.7 & 57.4 & {\color{gray} 60.2} & {\color{gray} 61.0} \\
~  &  FSOD-KD~(2022)  &  Faster R-CNN~(R-101)  & 39.3 & 44.4 & 47.5 & 52.8 & 54.2 \\
~  &  FADI~(2022)  &  Faster R-CNN~(R-101)  & 42.2 & 46.5 & 47.9 & 52.4 & 56.9 \\
~  &  PSEUDO~(2022)  &  Faster R-CNN~(R-101)  & 45.2 & 45.0 & 54.8 & 57.5 & 58.6 \\
~  &  FSOD-DIS~(2022)  &  Faster R-CNN~(R-101)  & \textbf{\color{red} 53.9} & \textbf{\color{red} 57.0} & \textbf{\color{red} 60.0} & \textbf{\color{red} 62.3} & \textbf{\color{red} 61.9} \\

\bottomrule
\end{tabular}}
\end{table}

%% file: n07_directions.tex
\section{Promising Directions\label{sec:future_dirs}}

\subsection{Promising Directions for FSOD}

Since FSOD extends OSOL by withdrawing the prior information of test images, this survey discusses the promising directions of FSOD to provide guidance for both FSOD and OSOL.

\subsubsection{\textbf{Efficient FSOD}}

FSOD models are generally modified from representative object detectors like Faster R-CNN, YOLO-style detectors. Current FSOD methods need to first pre-train these models on the data-abundant base dataset, then fine-tune them on the data-scarce novel dataset.
The pre-training on the base dataset requires a large device cost and time cost similar to general object detection. Besides, current methods spend much time during the few-shot fine-tuning stage for the model to converge~(usually more than $10$ epochs). The high computing cost of the model and long convergence time prevent FSOD from the real-life application. Therefore, lightweight and quickly-converged methods are required for future FSOD.

\input{tables/fsod_coco_benchmark}

\input{tables/zsod_benchmark_4817}
\input{tables/zsod_benchmark_6515}



\subsubsection{\textbf{Cross-Domain FSOD}}

Almost all of the current FSOD methods are evaluated in the single-domain condition.
Cross-domain few-shot learning is a more realistic setting that the data for base classes and novel classes are drawn from two domains.
Some studies \cite{Guo2020Broader} on cross-domain few-shot image classification indicate that the few-shot method does not have consistent performance in the single-domain condition and cross-domain condition. For example, this paper demonstrates that although some meta-learning methods achieve better performance than fine-tuning methods in the single-domain condition, they significantly underperform even some simple fine-tuning methods in the cross-domain condition.
Cross-domain few-shot object detection is a more complicated task than cross-domain few-shot image classification.
Recently a few methods~\cite{gao2022ECCV, lee2022ECCV, xiong2022cd} propose some benchmarks on cross-domain FSOD and set up some baselines for this area. Nevertheless, cross-domain FSOD deserves more exploration in the future for its practicality.

\subsubsection{\textbf{New Detection Framework for FSOD}}

Most of the current FSOD methods adopt Faster R-CNN as the detection framework. Some other powerful frameworks are worth exploring in the future. For example, vision transformer focuses more on holistic information of the image than local information, which can capture features missed by traditional CNN models. Currently, it has been widely applied in many other computer vision areas. In FSOD, the recently proposed Meta-DETR has improved the performance of FSOD to the SOTA on the MS COCO benchmark, which exceeds previous Faster R-CNN based detectors by several points. Therefore, the potential of vision transformer on FSOD still requires exploration.

\subsection{Promising Directions for ZSOD}

\subsubsection{\textbf{Combining Auxiliary Information for ZSOD}}

Combining information from an external source to assist ZSOD is a potential direction for performance improvement.
Some ZSOD methods attempt to exploit the information of external classes (not intersecting with base classes and novel classes) to augment semantic attributes of base classes and novel classes.
Moreover, some other ZSOD methods utilize an external word vocabulary to enhance the visual-semantic mapping.
However, no ZSOD method delves into the utilization of external auxiliary information as a whole, which requires more attention in the future.



\subsubsection{\textbf{Large Cross-Modal Foundation Model for ZSOD}}
Recently some large pre-trained cross-modal model show incredibly strong performance in aligning the context semantic between images and their text descriptions.
CLIP~\cite{Radford2021ICML} is the representative work of these large cross-modal models. Specifically, CLIP pre-trains the model on a large-scale dataset comprising abundant image-text pairs. CLIP encodes the images and texts with two parallel transformer-based models and adopts a contrastive learning strategy for training.
CLIP has the strong capacity of projecting images and texts into a common feature space, thus it can be directly transferred to the zero-shot scenario.
Recently, CLIP has been widely adopted for open-vocabulary object detection.

\subsubsection{\textbf{ZSOD combined with FSOD}}

A more generic scenario may appear in real-life where only some novel classes have annotated samples, yet other novel classes have semantic attributes, which requires the combination of ZSOD and FSOD. Some methods have been proposed to tackle this scenario. For example, ASD \cite{Shafin2020Anyshot} and UniT \cite{siddhesh2021unit} introduce an LSOD setting that the model makes predictions utilizing both semantic information and image samples. Moreover, UniT significantly improves the performance of FSOD with auxiliary semantic information.
Therefore, this generalized setting has more practical significance for the application of LSOD in the future.

%% file: tables/fsod_coco_benchmark.tex
\begin{table}
\footnotesize
\renewcommand\arraystretch{1.1}
\centering
\caption{Performance (mAP) of FSOD methods on the MS COCO benchmark~(only the methods with released codes are listed). These FSOD methods are evaluated under the $1, 2, 3, 5, 10, 30$-shot conditions. For each shot, the red font denotes the best performance, and the gray font denotes the second-best performance. ${}^{\star}$ denotes that the results are averaged over multiple runs, and R-50 \& R-101 denote ResNet-50 \& ResNet-101.}
\label{tab:coco_performance}
\setlength{\tabcolsep}{1.2mm}{
\begin{tabular}{c|c|c| 
*6{c}} 
    \toprule
  ~ & \textbf{Method} & \textbf{Backbone} & \textbf{1} & \textbf{2} & \textbf{3} & \textbf{5} & \textbf{10} & \textbf{30} \\
  \midrule
  \multirow{13}*{\rotatebox{270}{\textbf{\small Meta-Learning}}} & FSRW~(2018) & YOLOv2 & N/A & N/A & N/A & N/A & 5.6 & 9.1 \\

~ & Meta-RCNN~(2019) & Faster R-CNN~(R-101) & N/A & N/A & N/A & N/A & 8.7 & 12.4 \\

~ & FsDet~(2020)${}^{\star}$ & Faster R-CNN~(R-101) & 4.5 & 6.6 & 7.2 & 10.7 & 12.5 & 14.7 \\

~ & Attention-RPN~(2020) & Faster R-CNN~(R-50) & 4.2 & 5.6 & 6.6 & 8.0 & 11.1 & 13.5 \\

~ & DRL~(2021)${}^{\star}$ & Faster R-CNN~(R-101) & N/A & N/A & N/A & N/A & 11.9 & 14.6 \\

~ & DCNet~(2021)${}^{\star}$ & Faster R-CNN~(R-101) & N/A & N/A & N/A & N/A & 12.8 & 18.6 \\

~ & CME~(2021) & YOLOv2 & N/A & N/A & N/A & N/A & 15.1 & 16.9 \\

~ & Meta-DETR~(2022)${}^{\star}$ & Def. DETR~(R-101) & {\color{gray} 7.5} & N/A & {\color{gray} 13.5} & {\color{gray} 15.4} & 19.0 & 22.2 \\

~ & QA-FewDet~(2021) & Faster R-CNN~(R-101) & 4.9 & 7.6 & 8.4 & 9.7 & 11.6 & 16.5 \\

~ & DAnA-FasterRCNN~(2021) & Faster R-CNN~(R-50) & N/A & N/A & N/A & N/A & 18.6 & 21.6 \\

~ & Meta Faster R-CNN~(2022) & Faster R-CNN~(R-101) & 5.1 & 7.6 & 9.8 & 10.8 & 12.7 & 16.6 \\

~ & FCT~(2022)${}^{\star}$ & Faster R-CNN~(PVTv2-B2-Li) & 5.1 & 7.2 & 9.8 & 12.0 & 15.3 & 20.2 \\

~ & VFA~(2023) & Faster R-CNN~(R-101) & N/A & N/A & N/A & N/A & 16.2 & 18.9 \\

\hline

\multirow{14}*{\rotatebox{270}{\textbf{\small Transfer-Learning}}} & TFA w/cos~(2020) & Faster R-CNN~(R-101) & 3.4 & 4.6 & 6.6 & 8.3 & 10.0 & 13.7 \\

~ & Halluc~(2021) & Faster R-CNN~(R-101) & 4.4 & 5.6 & 7.2 & N/A & N/A & N/A \\

~ & MPSR~(2020) & Faster R-CNN~(R-101) & 2.3 & 3.5 & 5.2 & 6.7 & 9.8 & 14.1 \\


~ & ${\rm FSOD}^{\rm up}~(2021)$ & Faster R-CNN~(R-101) & N/A & N/A & N/A & N/A & 11.0 & 15.6 \\

~ & FSCE~(2021)${}^{\star}$ & Faster R-CNN~(R-101) & N/A & N/A & N/A & N/A & 11.9 & 16.4 \\

~ & DeFRCN~(2021)${}^{\star}$ & Faster R-CNN~(R-101) & 4.8 & {\color{gray} 8.5} & 10.7 & 13.6 & 16.8 & 21.2 \\

~ & N-PME~(2022) & Faster R-CNN~(R-101) & N/A & N/A & N/A & N/A & 10.6 & 14.1 \\

~ & FSOD-KI~(2022) & Faster R-CNN~(R-101) & N/A & N/A & N/A & N/A & 13.0 & 16.8 \\

~ & FSOD-KD~(2022) & Faster R-CNN~(R-101) & N/A & N/A & N/A & N/A & 12.5 & 17.1 \\

~ & FADI~(2022) & Faster R-CNN~(R-101) & N/A & N/A & N/A & N/A & 12.2 & 16.1 \\

~ & PSEUDO~(2022) & Faster R-CNN~(Swin-S) & N/A & N/A & N/A & N/A & 19.0 & 26.8 \\

~ & FSOD-DIS~(2022) & Faster R-CNN~(R-101) & \textbf{\color{red} 10.8} & \textbf{\color{red} 13.9} & \textbf{\color{red} 15.0} & \textbf{\color{red} 16.4} & 19.4 & 22.7 \\

~ & imTED~(2022) & Faster R-CNN~(ViT-B) & N/A & N/A & N/A & N/A & {\color{gray} 22.5} & \textbf{\color{red} 30.2} \\

~ & DETReg~(2022) & Def. DETR~(R-50) & N/A & N/A & N/A & N/A & \textbf{\color{red} 25.0} & {\color{gray} 30.0} \\
  \bottomrule
  \end{tabular}}
\end{table}

%% file: tables/zsod_benchmark_4817.tex
\begin{table*}
\footnotesize
\renewcommand\arraystretch{1.2}
\centering
\caption{Performance (mAP50) of ZSOD methods on the MS COCO Benchmark~(Seen classes/Unseen classes = \textbf{48/17}). \textbf{ZSOD} denotes the performance under ZSOD protocol. \textbf{Seen}, \textbf{Unseen} and \textbf{HM} denote the performance of base classes, novel classes and their harmonic average under GZSOD protocol, respectively. For each column, the red font denotes the best performance, and the gray font denotes the second-best performance. R-50 \& R-101 denote ResNet-50 \& ResNet-101.}
\label{tab:gzsd_coco_benchmark_4817}
\setlength{\tabcolsep}{0.35mm}{
\begin{tabular}{c|c|c|*2{c}|*2{c}|*2{c}|*2{c}} 
    \toprule
  \multirow{2}*{\textbf{Method}} & \multirow{2}*{\textbf{Semantic}} & \multirow{2}*{\textbf{Detector~(Backbone)}} & \multicolumn{2}{c|}{\textbf{ZSOD}} & \multicolumn{2}{c|}{\textbf{Seen}} & \multicolumn{2}{c|}{\textbf{Unseen}} & \multicolumn{2}{c}{\textbf{HM}} \\
  
  ~ & ~ & ~ & mAP & Recall & mAP & Recall & mAP & Recall & mAP & Recall \\
  \midrule
  SB~(2018)~\cite{bansal2018zsd} & GloVe & Faster R-CNN~(Inception) & 0.70 & 24.39 & N/A & N/A & N/A & N/A & N/A & N/A \\
  DSES~(2018)~\cite{bansal2018zsd} & GloVe & Faster R-CNN~(Inception) & 0.54 & 27.19 & N/A & 15.02 & N/A & 15.32 & N/A & 15.17 \\
  TOPM~(2019)~\cite{shao2019zero} & GloVe & YOLOv3~(DarkNet-53) & \textbf{\color{red} 15.43} & 39.20 & N/A & N/A & N/A & N/A & N/A & N/A \\
  CG-ZSOD~(2020)~\cite{li2020context} & BERT & YOLOv3~(DarkNet-53) & 7.20 & N/A & N/A & N/A & N/A & N/A & N/A & N/A \\
  GTNet~(2020)~\cite{zhao2020gtnet} & fastText & Faster R-CNN~(R-101) & N/A & 44.6 & N/A & N/A & N/A & N/A & N/A & N/A \\
  JRLNC-ZSOD~(2020)~\cite{rahman2020Joint} & word2vec & Faster R-CNN~(R-50) & 5.05 & 12.27 & 13.93 & 20.42 & 2.55 & 12.42 & 4.31 & 15.45 \\
  SPGP~(2020)~\cite{caixia2020semantics} & word2vec & Faster R-CNN~(R-101) & N/A & 35.40 & N/A & N/A & N/A & N/A & N/A & N/A \\
  VSA-ZSOD~(2020)~\cite{rahman2020visual_semantic} & word2vec & RetinaNet~(R-50) & 10.01 & 43.56 & 35.92 & 38.24 & 4.12 & 26.32 & 7.39 & 31.18 \\
  MS-Zero++~(2020)~\cite{gupta2020multispace} & word2vec & Faster R-CNN~(R-101) & N/A & N/A & 35.00 & N/A & \textbf{\color{red} 13.80} & 35.00 & {\color{gray} 19.80} & N/A \\
  BLC~(2020)~\cite{ye2020background} & word2vec & Faster R-CNN~(R-50) & 10.60 & 48.87 & 42.10 & 57.56 & 4.50 & 46.39 & 8.20 & 51.37 \\
  ZSI~(2021)~\cite{ye2021seg} & word2vec & Faster R-CNN~(R-101) & 11.40 & 53.90 & {\color{gray} 46.51} & {\color{gray} 70.76} & 4.83 & 53.85 & 8.75 & \textbf{\color{red} 61.16} \\
  ZSDTR~(2021)~\cite{zheng2021ICIP} & word2vec & Def. DETR~(R-50) & 10.40 & 48.50 & \textbf{\color{red} 48.53} & \textbf{\color{red} 74.31} & 5.62 & 48.44 & 9.45 & {\color{gray} 60.53} \\
  VSRG~(2022)~\cite{nie2022WACV} & word2vec & Faster R-CNN~(R-50) & 11.40 & {\color{gray} 55.03} & 43.90 & 66.70 & 4.70 & {\color{gray} 54.54} & 8.50 & 60.01 \\
  ContrastZSOD~(2022)~\cite{yan2022semantics} & word2vec & Faster R-CNN~(R-101) & 12.50 & 52.40 & 45.10 & 65.70 & 6.30 & 52.40 & 11.10 & 58.30 \\
  RRFS-ZSOD~(2022)~\cite{huang2022CVPR}) & fastText & Faster R-CNN~(R-101) & {\color{gray} 13.40} & 53.50 & 42.30 & 59.70 & 13.40 & \textbf{\color{red} 58.80} & \textbf{\color{red} 20.40} & 59.20 \\
  CLIP-ZSOD~(2022)~\cite{xie2022ICDM} & word2vec & YOLOv5~(CSPDarkNet-53) & {\color{gray} 13.40} & \textbf{\color{red} 55.80} & 31.70 & 63.30 & {\color{gray} 13.60} & 45.20 & 19.00 & 52.70 \\
  
  \bottomrule
  \end{tabular}}
\end{table*}

%% file: tables/zsod_benchmark_6515.tex
\begin{table*}
\footnotesize
\renewcommand\arraystretch{1.2}
\centering
\caption{Performance (mAP50) of ZSOD methods on the MS COCO Benchmark~(Seen classes/Unseen classes = \textbf{65/15}). \textbf{ZSOD} denotes the performance under ZSOD protocol. \textbf{Seen}, \textbf{Unseen} and \textbf{HM} denote the performance of base classes, novel classes and their harmonic average under GZSOD protocol, respectively. For each column, red font denotes the best performance, and gray font denotes the second-best performance.}
\label{tab:gzsd_coco_benchmark_6515}
\setlength{\tabcolsep}{0.65mm}{
\begin{tabular}{c|c|c|*2{c}|*2{c}|*2{c}|*2{c}} 
    \toprule
  \multirow{2}*{\textbf{Method}} & \multirow{2}*{\textbf{Semantic}} & \multirow{2}*{\textbf{Detector~(Backbone)}} & \multicolumn{2}{c|}{\textbf{ZSOD}} & \multicolumn{2}{c|}{\textbf{Seen}} & \multicolumn{2}{c|}{\textbf{Unseen}} & \multicolumn{2}{c}{\textbf{HM}} \\
  
  ~ & ~ & ~ & mAP & Recall & mAP & Recall & mAP & Recall & mAP & Recall \\
  \midrule
  
  Transductive~(2019)~\cite{rahman2019transductive} & word2vec & RetinaNet~(R-50) & 14.57 & 48.15 & 28.78 & 54.14 & 14.05 & 37.16 & 18.89 & 44.07 \\
  CG-ZSOD~(2020)~\cite{li2020context} & BERT & YOLOv3~(DarkNet-53) & 10.90 & N/A & N/A & N/A & N/A & N/A & N/A & N/A \\
  LSA-ZSOD~(2020)~\cite{wang2020learning} & aPaY & RetinaNet~(R-50) & 13.55 & 37.78 & 34.18 & 40.32 & 13.42 & 38.73 & 19.27 & 39.51 \\
  ACS-ZSOD~(2020)~\cite{mao2020zero} & aPaY & RetinaNet~(R-50) & 15.34 & 47.83 & N/A & N/A & N/A & N/A & N/A & N/A \\
  SYN-ZSOD~(2020)~\cite{hayat2020synthesizing} & fastText & Faster R-CNN~(R-101) & 19.00 & 54.00 & 36.90 & 57.70 & 19.00 & 53.90 & 25.08 & 55.74 \\
  VSA-ZSOD~(2020)~\cite{rahman2020visual_semantic} & word2vec & RetinaNet~(R-50) & 12.40 & 37.72 & 34.07 & 36.38 & 12.40 & 37.16 & 18.18 & 36.76 \\
  BLC~(2020)~\cite{ye2020background} & word2vec & Faster R-CNN~(R-50) & 14.70 & 54.68 & 36.00 & 56.39 & 13.10 & 51.65 & 19.20 & 53.92 \\
  ZSI~(2021)~\cite{ye2021seg} & word2vec & Faster R-CNN~(R-101) & 13.60 & 58.90 & 38.68 & {\color{gray} 67.11} & 13.60 & 58.93 & 20.13 & 62.76 \\
  ZSDTR~(2021)~\cite{zheng2021ICIP} & word2vec & Def. DETR~(R-50) & 13.20 & 60.30 & \textbf{\color{red} 40.55} & \textbf{\color{red} 69.12} & 13.22 & 59.45 & 20.16 & 61.12 \\
  DPIF~(2021)~\cite{li2021inference} & word2vec & Faster R-CNN~(R-50) & 19.82 & 55.73 & 29.82 & 56.68 & 19.46 & 38.70 & 23.55 & 46.00 \\
  VSRG~(2022)~\cite{nie2022WACV} & word2vec & Faster R-CNN~(R-50) & 14.90 & 62.70 & 38.10 & 65.31 & 13.90 & 60.52 & 20.40 & {\color{gray} 62.82} \\
  RSC-ZSOD~(2022)~\cite{sarma2022resolving} & word2vec & Faster R-CNN~(R-101) & {\color{gray} 20.10} & {\color{gray} 65.10} & 37.40 & 58.60 & {\color{gray} 20.10} & {\color{gray} 64.00} & {\color{gray} 26.15} & 61.18 \\
  ContrastZSOD~(2022)~\cite{yan2022semantics} & word2vec & Faster R-CNN~(R-101) & 18.60 & 59.50 & {\color{gray} 40.20} & 62.90 & 16.50 & 58.60 & 23.40 & 60.70 \\
  RRFS-ZSOD~(2022)~\cite{huang2022CVPR}) & fastText & Faster R-CNN~(R-101) & 19.80 & 62.30 & 37.40 & 58.60 & 19.80 & 61.80 & 26.00 & 60.20 \\
  CLIP-ZSOD~(2022)~\cite{xie2022ICDM} & word2vec & YOLOv5~(CSPDarkNet-53) & 18.30 & \textbf{\color{red} 69.50} & 31.70 & 61.00 & 17.90 & \textbf{\color{red} 65.20} & 22.90 & \textbf{\color{red} 63.00} \\
  CCFA-ZSOD~(2022)~\cite{li2022zero} & word2vec & RetinaNet~(R-50) & \textbf{\color{red} 24.62} & 55.32 & 33.35 & 38.64 & \textbf{\color{red} 24.62} & 54.72 & \textbf{\color{red} 28.31} & 45.29 \\
  
  \bottomrule
  \end{tabular}}
\end{table*}

%% file: n08_conclusions.tex
\section{Conclusion\label{sec:conclusion}}

Enhancing the deep object detectors to quickly learn from very few or even zero samples is of great significance to future object detection.
This paper conducts a comprehensive survey on Low-Shot Object Detection~(LSOD), consisting of One-Shot Object Localization~(OSOL), Few-Shot Object Detection~(FSOD) and Zero-Shot Object Detection~(ZSOD).
In this survey, the emergence background and evolution history of LSOD are first reviewed. Then, current LSOD methods are analyzed systematically based on an explicit and complete taxonomy of these methods, including some extensional topics of LSOD. Moreover, the pros and cons of LSOD methods are indicated with a comparison of their performance. Finally, the challenges and promising directions of LSOD are discussed. Hopefully, this survey can promote future research on LSOD.